\documentclass[acmlarge]{acmart}
\AtBeginDocument{%
  }

\setcopyright{acmlicensed}
\copyrightyear{2018}
\acmYear{2018}
\acmDOI{XXXXXXX.XXXXXXX}

\acmJournal{IMWUT}
\acmYear{2024} \acmVolume{1} \acmNumber{1} \acmArticle{1} \acmMonth{1} \acmPrice{15.00}\acmDOI{10.1145/3459666}

\usepackage{multirow,multicol}
\usepackage[capitalise]{cleveref}
\usepackage{soul}

\begin{document}

\title{TxP: Reciprocal Generation of Ground Pressure Dynamics and Activity Descriptions for Improving Human Activity Recognition}

\author{Lala Shakti Swarup Ray}
\orcid{0000-0002-7133-0205}
\affiliation{%
  \institution{DFKI}
  \city{Kaiserslautern}
  \country{Germany}}
\email{lala_shakti_swarup.ray@dfki.de}

\author{Lars Krupp}
\orcid{0000-0001-6294-2915}
\affiliation{%
  \institution{RPTU and DFKI}
  \city{Kaiserslautern}
  \country{Germany}}
\email{lars.krupp@dfki.de}

\author{Vitor Fortes Rey}
\orcid{0000-0002-8371-2921}
\affiliation{%
  \institution{RPTU and DFKI}
  \city{Kaiserslautern}
  \country{Germany}}
\email{fortes@dfki.uni-kl.de}

\author{Bo Zhou}
\email{bo.zhou@dfki.de}
\orcid{0000-0002-8976-5960}
\affiliation{%
  \institution{RPTU and DFKI}
  \city{Kaiserslautern}
  \country{Germany}
}

\author{Sungho Suh}
\orcid{0000-0003-3723-1980}
\email{sungho_suh@korea.ac.kr}
\authornote{Corresponding author}
\affiliation{%
  \institution{Korea University}
  \city{Seoul}
  \country{Republic of Korea}
}

\author{Paul Lukowicz}
\orcid{0000-0003-0320-6656}
\affiliation{%
  \institution{RPTU and DFKI}
  \city{Kaiserslautern}
  \country{Germany}}
\email{Paul.Lukowicz@dfki.de}

\renewcommand{\shortauthors}{Ray et al.}

\begin{abstract}
Sensor-based human activity recognition (HAR) has predominantly focused on Inertial Measurement Units and vision data, often overlooking the capabilities unique to pressure sensors, which capture subtle body dynamics and shifts in the center of mass.
Despite their potential for postural and balance-based activities, pressure sensors remain underutilized in the HAR domain due to limited datasets. 
To bridge this gap, we propose to exploit generative foundation models with pressure-specific HAR techniques. 
Specifically, we present a bidirectional Text$\times$Pressure model that uses generative foundation models to interpret pressure data as natural language. 
TxP accomplishes two tasks: (1) Text2Pressure, converting activity text descriptions into pressure sequences, and (2) Pressure2Text, generating activity descriptions and classifications from dynamic pressure maps. 
Leveraging pre-trained models like CLIP and LLaMA 2 13B Chat, TxP is trained on our synthetic PressLang dataset, containing over 81,100 text-pressure pairs. 
Validated on real-world data for activities such as yoga and daily tasks, TxP provides novel approaches to data augmentation and classification grounded in atomic actions.
This consequently improved HAR performance by up to 12.4\% in macro F1 score compared to the state-of-the-art, advancing pressure-based HAR with broader applications and deeper insights into human movement. The data and code will be available on \href{https://github.com/lalasray/TxP}{TxP}.
\end{abstract}




\maketitle
\section{Introduction}

Human Activity Recognition (HAR) has emerged as a pivotal area of research \cite{gupta2022human} aimed at understanding and classifying human movements across diverse modalities, integral to applications such as healthcare monitoring \cite{soni2023novel, islam2023multi}, sports performance analysis \cite{afsar2023body, xiao2023recognizing}, and smart home automation \cite{diraco2023review}. 
By enabling systems to recognize and interpret human actions, HAR enhances the functionality of these applications \cite{snoun2023deep}, making them more adaptive to users' needs. 
Traditionally, vision-based modalities have been the dominant approach in HAR \cite{bhola2024review}, relying on external cameras and image pre-processing techniques to capture human motion, which is slow and computationally intensive.
In contrast, sensor-based HAR \cite{ni2024survey, ankalaki2024simple} has gained traction due to its ability to capture real-time data directly from the body, offering key advantages in privacy, portability, and adaptability across environments. 
Unlike vision systems that require constant visual input and can be intrusive, sensor-based data is not rich enough for person identification, thereby addressing privacy concerns \cite{ray2024har}.
Sensor-based HAR systems have traditionally relied on Inertial Measurement Units (IMUs) \cite{bello2023captainglove,liu2024imove} and Electromyography (EMG) sensors \cite{rani2023surface, vamsi2024efficient}, which capture movement changes and muscle activations, respectively. 
In contrast, pressure sensors \cite{singh2024novel, zhou2022quali} provide unique insights by recording body-ground interactions. These sensors can capture fine details such as subtle shifts in pressure distribution, the center of mass, and force application \cite{scott2020image}, which are essential for analyzing balance, posture, and gait.
This level of detail is crucial for specific activities where analyzing balance, posture, and gait is essential, such as gait analysis \cite{jeon2020novel}, rehabilitation \cite{liu2023aloe}, sports performance optimization \cite{jeong2021ultra}, and fall risk assessment for the elderly \cite{bucinskas2021wearable}, offering insights that cannot be derived from IMUs or EMGs alone.
These are particularly effective for fine-grained analysis of foot-ground interactions \cite{zhong2021wide}, such as detecting irregularities in gait or assessing load distribution. 
Additionally, in scenarios where subtle shifts in pressure provide more valuable information than gross movements, such as balance training \cite{amit2020flexible} or prosthetic design \cite{tabor2021textile}, pressure sensors are indispensable.

However, despite their potential, pressure sensor-based HAR faces a significant barrier: the scarcity of large-scale public datasets. Collecting pressure data is expensive, time-consuming, and subject to privacy concerns, as the collection of personal data can raise ethical issues. The lack of publicly available datasets hinders the development of robust machine learning models, limiting their ability to generalize across different tasks.
Previous works have tried to simulate pressure signals from other modalities like 3D  pose or depth data using deep learning and 3D physics simulations\cite{ray2023pressim, clever2022bodypressure, scott2020image}.
However, most of them suffer from obvious limitations, such as being computationally expensive and slow, and having limited applications to the activity types they are trained on. 
Users also need to source the seed modalities and manually annotate them, which is a slow and labor-intensive process.

\begin{figure}[!t]
\centering
\includegraphics[width=0.8\textwidth]{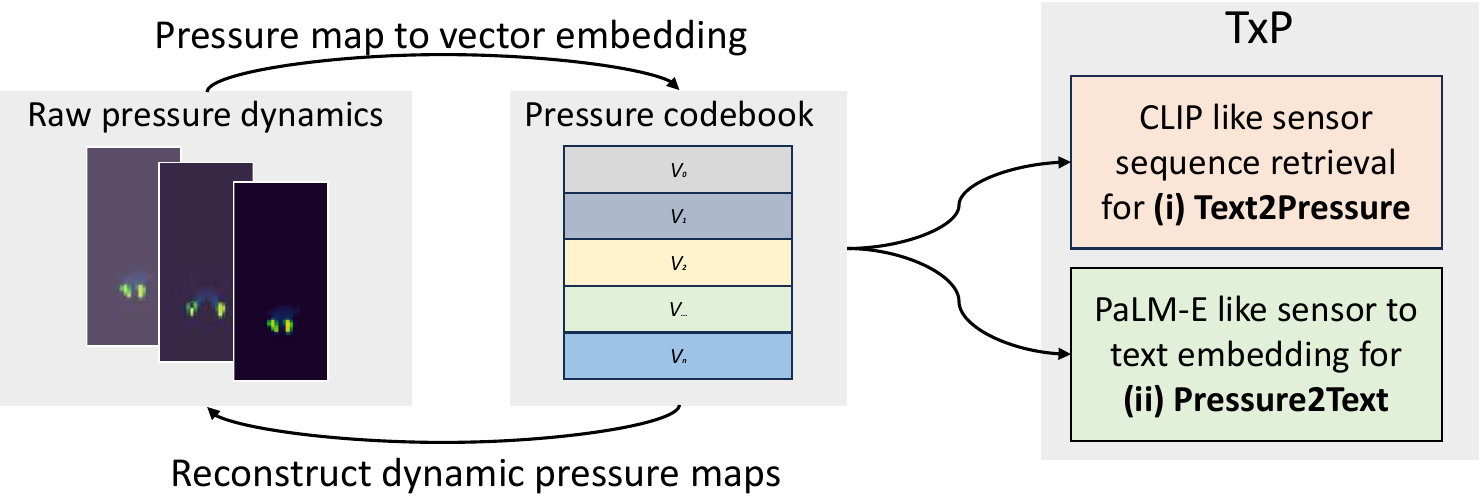} 
\caption{The TxP framework consists of two key components: (i) Text2Pressure, which generates dynamic pressure sequences from textual descriptions similar to CLIP \cite{radford2021learning}, and (ii) Pressure2Text, which creates activity descriptions from dynamic pressure sequences similar to PaLM-E \cite{driess2023palm}. At its core is PressureRQVAE, which creates a mapping between raw pressure dynamics and a Pressure codebook that contains pressure embeddings represented in a discrete vector space.} 
\label{overview}
\end{figure}

To address these challenges, we propose the Text$\times$Pressure (TxP) framework, which facilitates the synthesis of large, high-quality pressure datasets from textual descriptions of human activities and generates open-ended text from dynamic pressure sequences as visualized in \cref{overview}. By leveraging cross-modal learning techniques, TxP enables the generation of synthetic pressure data that mimics real-world interactions without needing slow and cumbersome data collection. This approach alleviates the limitation of data scarcity without raising any privacy concerns by reducing reliance on personal data collection. Creating a model that generates atomic-level activity descriptions from a dynamic sequence of pressure maps helps further enhance pressure-based HAR systems.

Moreover, our extensive experiments on four real-world datasets demonstrate that synthetic data generation through TxP enhances model performance by providing diverse and well-structured training data that mirrors real-world complexity. By bridging the gap between text and pressure data, TxP allows the creation of datasets that retain the statistical properties of actual sensor readings, enabling effective model training without compromising user privacy.
In addition to generating synthetic data, the TxP framework introduces a novel approach to transforming pressure sequences into atomic-level activity descriptions while allowing dynamic input lengths to enhance the interpretability of HAR models.
This bidirectional capability opens new pathways for understanding and analyzing human activities, making HAR models more adaptable and applicable to real-world scenarios.

Our contributions can be summarized as follows:

\begin{itemize}
    \item \textbf{TxP Framework}: We introduce a novel \textit{TxP} framework that facilitates the reciprocal generation of synthetic pressure data and textual descriptions of activities. The \textit{Text2Pressure} component generates high-quality synthetic pressure sequences from natural language descriptions, augmenting the training of HAR models. The \textit{Pressure2Text} component improves model interpretability by producing detailed, natural language descriptions of pressure sequences, providing context beyond simple classification labels. This bidirectional framework addresses the challenges of data scarcity and enhances HAR models' adaptability to diverse real-world applications.

    \item \textbf{PressLangDataset}: We present a large-scale synthetic dataset containing paired pressure sequences and natural language descriptions of the physical activity. The dataset offers a comprehensive resource for training pressure-based models with 81.1K unique motions and 78 million individual pressure frames. PressLangDataset captures various activities across various body types, providing diverse and representative data for pressure-based HAR research.

    \item \textbf{HAR Improvement}: We conduct extensive qualitative and quantitative evaluations of both the Text2Pressure and Pressure2Text components, using Pressim \cite{ray2023pressim}, TMD \cite{ray2024text}, PmatData \cite{pouyan2017pressure}, MeX \cite{wijekoon2019mex} and PID4TC \cite{o2024ai} dataset to assess the effectiveness of our framework and conclude that Text2Pressure data augmentation alone can improve HAR results by up to 11.6$\%$, and Pressure2Text classification grounded with atomic actions yields an improvement of up to 8.2$\%$. Their combined use results in a macro F1 score gain of up to 12.4$\%$, surpassing the current state-of-the-art on pressure-based HAR datasets.
    
\end{itemize}

The rest of the paper is organized as follows: \cref{sec:2} reviews related work in pressure-based HAR, synthetic data generation, and cross-modal understanding. \cref{sec:3} provides a detailed description of the TxP framework, including the construction of the PressLangDataset and the development of the PressureRQVAE, Text2Pressure, and Pressure2Text components. \cref{sec:4} outlines our evaluation methodology, results along with qualitative and quantitative comparison, \cref{sec:discussion} discusses their implications and current limitations, and \cref{sec:5} concludes the paper with a discussion of future directions.

\section{Related Work}
\label{sec:2}

\subsection{Pressure-Based Human Motion Understanding}

Recent advancements in pressure-based human motion understanding have enabled a wide range of applications in activity recognition. For instance, smart shoes with embedded pressure sensors support daily activity monitoring \cite{truong2019wearable}, and pressure-sensitive mattresses are used for recognizing body-weight exercises \cite{zhou2022quali}. In construction, pressure insoles assist in tracking worker activities \cite{antwi2020construction}, while in sports and fitness, devices like SmarCyPad—a pressure sensor-based seat—enhance cycling performance monitoring \cite{wu2023smarcypad}. 

Wearable pressure-sensing insoles have emerged as a lightweight yet effective solution for capturing foot-ground interaction patterns essential for both fine-grained activity recognition \cite{d2022assessing} and full-body pose estimation \cite{wu2024soleposer, hansmart}, particularly when combined with deep learning models trained on human motion dynamics. These systems offer an unobtrusive alternative to vision-based or full-body sensor setups. In addition, pressure sensor gateways contribute to gait analysis \cite{sanders2024concurrent}, and wearable pressure sensors are also employed in physiological monitoring, such as pulse wave tracking \cite{meng2022wearable}.

In smart home environments, pressure sensors play a growing role in elderly care applications \cite{gnanavel2016smart}. They are further used in health diagnostics, including pressure-sensor-based face masks for detecting abnormal breathing patterns \cite{zhong2022smart}, insoles for identifying Parkinson’s disease symptoms \cite{marcante2020foot}, and skin-mounted sensors for hand rehabilitation \cite{han2022smart}. Pressure-sensitive mattresses have also shown promise in 3D pose estimation tasks \cite{chen2024cavatar, wu2024seeing}.

Despite their versatility, conventional pressure-based human activity recognition (HAR) systems typically rely on discrete classification and fixed-length input windows, which restrict generalization to unseen classes. To address these limitations, we propose \textit{Pressure2Text}, a classifier that integrates a large language model (LLM) to process variable-length sequences and generate natural language descriptions of atomic activities. These descriptions are then mapped to activity classes using prompt engineering, allowing for a more flexible and generalizable approach to pressure-based motion understanding.

\subsection{Synthetic Sensor Data Generation for Improving HAR}

Numerous techniques have been explored to generate synthetic sensor data for enhancing HAR, each with distinct trade-offs. Basic approaches use neural networks to map one modality to another. For example, SDA-SPS \cite{zolfaghari2024sensor} generates IMU data from 3D pose sequences, and PressNet \cite{scott2020image} produces pressure maps from 2D pose snapshots. While these neural methods are quick and straightforward to train, they struggle with generalization and often fail to represent full-body or complex activities.

More advanced techniques combine neural networks with 3D simulations. These approaches use physics-informed deep learning to generate sensor outputs from modalities like 3D poses. For instance, IMUTube and Multi3Net \cite{kwon2020imutube, fortes2024enhancing} use inverse kinematics with neural models to synthesize linear acceleration and angular velocity. Similarly, BodyPressureNet \cite{clever2022bodypressure} and Pressim \cite{ray2023pressim} combine simulation with learning to generate pressure maps from volumetric poses. Though more accurate, these methods are computationally intensive and require detailed simulation setups, making replication and scalability difficult.

Intra-modal generation techniques have emerged to mitigate these challenges. These include generative models and style transfer networks that augment data within the same modality. Examples include PressureTransferNet \cite{ray2023pressuretransfernet} for pressure data, EMG-GAN \cite{chen2022deep} for EMG signals, and SensoryGAN \cite{wang2018sensorygans} and BSDGAN \cite{hu2023bsdgan} for IMU data. These models benefit from already-labeled data, reducing annotation costs. 
However, they are constrained by the diversity of their source datasets and struggle to simulate out-of-distribution or novel activity patterns.

A newer direction involves generating sensor data directly from textual descriptions. For IMU data, Text-to-Motion models \cite{zhang2023generating} combined with inverse kinematics (e.g., IMUGPT \cite{leng2024imugpt}) have been used. For pressure data, TMD \cite{ray2024text} generates pressure maps using generative models trained on Text-to-Motion outputs. These direct methods offer faster inference by leveraging text annotations, but they require large corpora of paired data for effective training.

Building upon this, we introduce \textit{Text2Pressure}, which extends the TMD approach to a broader set of activities and considers individual characteristics such as body shape and gender. This results in more diverse and personalized synthetic data for pressure-based HAR.

\subsection{Sensor-Based Cross-Modal Foundational Models}

Cross-modal foundational models like CLIP \cite{radford2021learning} align vision and language by embedding both in a shared latent space using large-scale paired datasets. These models enable zero-shot transfer across modalities for various downstream tasks. Extensions such as PaLM-E \cite{driess2023palm} integrate sensory data into large language models for embodied control, while VideoLLaMA \cite{zhang2023video} and BLIP-2 \cite{li2023blip} advance multimodal understanding with temporal or vision-language alignment.

Despite success in vision-language tasks, integrating sensor data into such foundational models is still nascent. Some efforts—like Meta-Transformer \cite{zhang2023meta}, ImageBind \cite{girdhar2023imagebind}, and IMU2CLIP \cite{moon2023imu2clip}—have extended CLIP-like architectures to handle IMU, depth, and time-series data. Others, such as zero-shot HAR \cite{tong2021zero} and OneLLM \cite{han2024onellm}, attempt to unify multimodal signals, including IMU, for improved activity classification.

However, these models aim to learn joint representations for general-purpose tasks such as segmentation, forecasting, or prediction rather than focusing on pressure-based HAR. Moreover, no existing work aligns pressure sensor data with pre-trained LLMs.

To fill this gap, we propose \textit{Text$\times$Pressure (TxP)}, a cross-modal framework that integrates pressure sensor data with natural language representations. TxP enables bidirectional mappings, generating pressure sequences from text and producing textual descriptions from pressure data. TxP supports novel applications like natural-language-driven HAR, realistic synthetic dataset creation, and improved model interpretability by embedding pressure dynamics in a language-aligned space.

\section{Proposed Approach}
\label{sec:3}
This section presents the comprehensive approach for training TxP, rooted in NLP and cross-modal learning. Specifically, we establish a correspondence between the core building blocks of dynamic pressure profiles and vector representations within a codebook. Each discrete vector links directly to text concepts through a recursive mechanism, which utilizes a large synthetic dataset with paired text and pressure dynamics. This methodology comprises four key components, each essential to building the whole system:

This section presents the comprehensive approach for training TxP, rooted in NLP and cross-modal learning. Specifically, we establish a correspondence between the core building blocks of dynamic pressure profiles, two-dimensional spatiotemporal representations of pressure variations over time, also called pressure dynamics, and vector representations within a codebook. Each discrete vector links directly to text concepts through a recursive mechanism, which utilizes a large synthetic dataset with paired text and pressure dynamics. Pressure dynamics refer to how pressure distributions change over time, capturing subtle shifts in weight, movement, and body position. This methodology comprises four key components, each essential to building the whole system:
\begin{itemize} 
\item PressLang Dataset Creation: To accurately capture a wide range of pressure patterns with corresponding textual descriptions, we construct a large-scale, paired dataset of dynamic pressure and text data. 
This diversity is essential to train TxP to generalize across various pressure scenarios.

\item Reversible Dynamic Pressure Embedding Representations: We achieve efficient vector representations using PressureRQVAE, which creates a compact codebook. 
This component enables smooth conversion between pressure profiles and vector representations, ensuring the encoded pressure dynamics are limited in variety yet retain sufficient detail to reconstruct any pressure signal accurately. 

\item Text2Pressure Generation: We leverage CLIP's text embeddings to train a cross-modal autoregressive transformer, translating the text into codebook vectors. 
These vectors can then be reconstructed into dynamic pressure maps, providing a direct path from language to pressure data, which is crucial for generating pressure profiles from textual prompts.

\item Pressure2Text Inference: Finally, we map codebook vectors back to textual embeddings using a pre-trained LLM combined with a projection model. 
This allows TxP to generate descriptive text from pressure profiles, enabling a complete loop from text to pressure and back to text for interpretability and refinement.
\end{itemize}

\subsection{PressLang Dataset}
A substantial dataset covering various physical activities performed by subjects with diverse body types is essential to train a generative model like TxP. 
Here, we describe how we created such a dataset by combining extensive, existing motion-language datasets with 3D simulations and LLM-based prompting.

\paragraph{Motion-X Dataset}
Motion-X \cite{lin2024motion} is the largest motion language dataset meticulously designed to enhance the study of human motion by incorporating 3D whole-body SMPL-X \cite{pavlakos2019expressive} annotations, detailed semantic descriptions, and frame-level pose data. 
With a staggering 15.6 million 3D annotations and 81.1K sequence-level semantic descriptions, Motion-X is assembled from eight distinct sources that include established datasets like AMASS \cite{mahmood2019amass}, HAA500 \cite{chung2021haa500}, AIST \cite{tsuchida2019aist}, HuMMan \cite{cai2022humman}, GRAB \cite{taheri2020grab}, EgoBody \cite{zhang2022egobody}, BAUM \cite{zhalehpour2016baum}, Idea400 and curated online videos. 
It contains different categories of human activities, emphasizing body movements and addressing hand and facial expression elements often overlooked in prior datasets that predominantly focused on body motion alone.
We used motions from Motion-X as seeds to create the PressLang dataset.
 
\paragraph{Pressure Dynamics Simulation}
To train our model effectively, we require a substantial amount of data, which is why we generate a diverse range of synthetic body shapes from the 24 SMPL \cite{loper2023smpl} and 30 MANO \cite{romero2022embodied} components from the Motion-X dataset. 
Modifying the ten shape parameters allows us to create representations of various genders and body types. This is crucial for simulating how distinct physical characteristics like weight and height influence pressure dynamics during various activities.

We employ the PresSim framework  \cite{ray2023pressim}, which combines physics simulations with neural networks, to generate pressure profiles from SMPL pose sequences designed explicitly for a SensingTex pressure-sensing mattress with 80$\times$28 sensor array. 
While the PresSim framework is relatively slow and memory-intensive, it is the current state-of-the-art pressure sensor simulation model as compared to other models like Pressnet \cite{scott2020image} and Bodypressurenet \cite{clever2022bodypressure} and produces highly accurate pressure representations for different types of human motions, which is essential to train TxP. 
It has already been validated in PressureTransferNet \cite{ray2023pressuretransfernet} that can generate very realistic Pressure maps that can, in turn, improve HAR by 4.8\%.
We generate the pressure dynamics for each seed motion with five random body shape/gender variations.

\paragraph{Annotation for Text2Pressure}
Following the simulation of pressure dynamics, we undertake a comprehensive re-annotation phase with a strong focus on ground-body interactions to refine the Motion-X dataset. This process involves carefully removing non-essential components from the natural language annotations, retaining only information directly related to these interactions, and adding key physical characteristics like body shape, weight, and height for specific activities. For instance, consider the raw annotation:

\textit{"The man is smiling while jumping up and landing on both feet, showing a facial expression of joy and moving their arms energetically."}

To enhance the dataset's usability for training machine learning models, we employ prompt engineering with LLaMA 2 to refine the annotation. We craft a prompt instructing the model to focus on relevant details, filtering out unnecessary information while incorporating physical characteristics:

\textit{"Please refine the following sentence by removing non-essential information and keeping only the details relevant to the ground-body interaction during the described activity, along with physical characteristics of the body in terms of atomic motions. Input: 'The man is smiling while jumping up and landing on both feet, showing a facial expression of joy and moving their arms energetically.' Output:"}

The output from the model would be:

\textit{"A medium-build adult jumps up and lands on both feet."}

This streamlined approach improves the quality of the annotations and aligns the dataset more closely with our research goals. This alignment enables the development of models that can accurately predict pressure dynamics based on human activities and body characteristics. The processed annotations provide a clean, focused dataset crucial for training robust models capable of analyzing human motion in various contexts.

\subsection{Pressure Embedding and Quantization}
\begin{figure}[!t]
\centering
\includegraphics[width=1\textwidth]{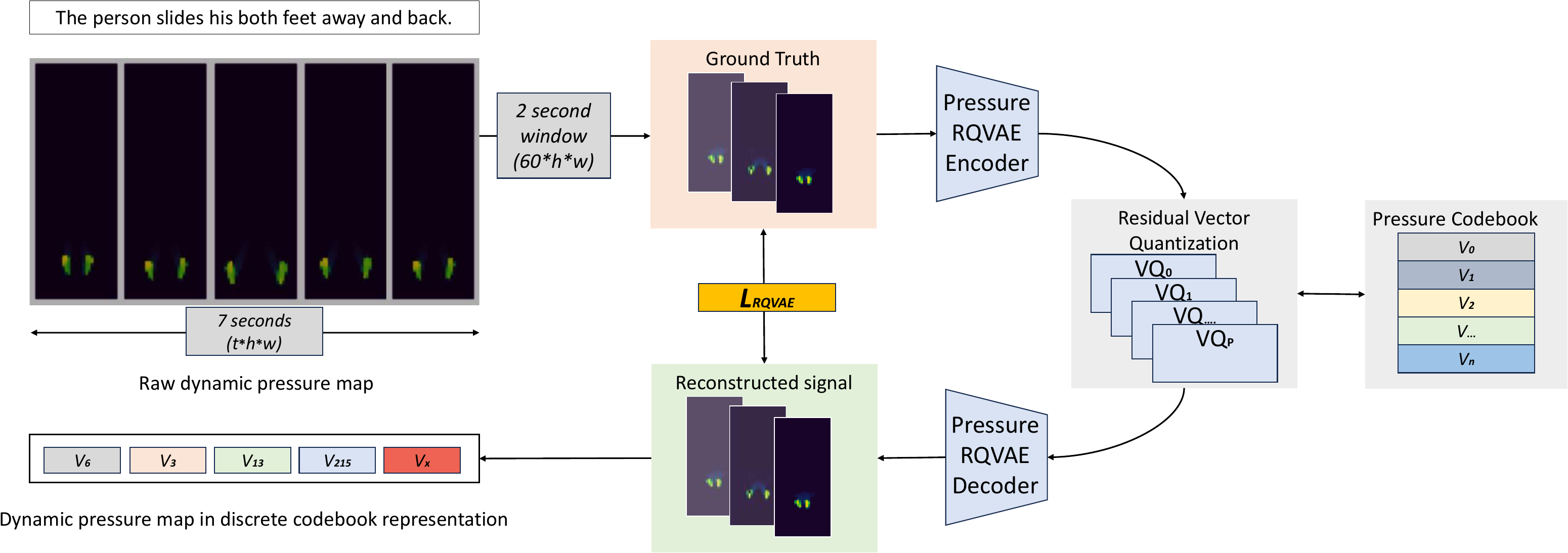} 
\caption{We used PressureRQVAE to quantize continuous dynamic length pressure time series data into discrete NLP-like tokens. The figure showcases one example of seven-second continuous pressure dynamics used to generate a discrete vector sequence of five tokens. Each token represents two seconds of pressure sequence data, the fourth token has only one second of pressure data and another second of null data (all pressure points have a null value), and finally, the fifth end token $V_x$ represents the end of the pressure sequence.} 
\label{PressureRQVAE}
\end{figure}

Pressure dynamics is a temporal sequence of pressure maps, and the goal is to create a system that can represent distinct spatial-temporal patterns and reconstruct them back into their original form. We use an autoencoder to encode the continuous pressure data into a lower-dimensional latent space to achieve this. However, the challenge is that traditional autoencoders create continuous latent spaces, which are difficult to map to discrete patterns. To address this, we employ the Residual Quantized Variational Autoencoder (RQVAE) \cite{lee2022autoregressive}, which refines the representation by quantizing the latent space across multiple layers. This results in a discrete set of tokens that can effectively represent the pressure dynamics while allowing for accurate reconstruction of the pressure maps. RQVAE ensures that the mapping process maintains the integrity of the original pressure data while making the system more compatible with discrete data processing systems like those used for text.

Dynamic pressure maps form a continuous time-series signal. 
At the same time, text data consists of a structured combination of discrete syntax and semantics. 
Both data streams need to be in a compatible format to create a system that maps between text and pressure data. 
We can achieve this by segmenting the continuous pressure signal into smaller chunks and creating embeddings for each segment using an autoencoder. This allows for the reconstruction of the pressure signal from its embedding. 
However, unlike text embeddings, which are limited to a fixed vocabulary, a simple autoencoder can generate a nearly unbounded range of pressure embeddings, complicating the task of mapping between the two modalities, as shown in TDM \cite{ray2024text}. 
To address this and reduce the variety of embeddings to a finite set, a vector-quantized variational autoencoder (VQ-VAE) can be employed.
This transformation is achieved through our Pressure Residual Quantized Variational Autoencoder (PressureRQVAE), an adaptation of the Residual Vector Quantized Variational Autoencoder (RQ-VQ-VAE), the current state-of-the-art VQVAE originally introduced in MoMask \cite{guo2024momask} for 3D pose quantization. 
PressureRQVAE converts continuous pressure patterns into tokenized sequences, as illustrated in \cref{PressureRQVAE}. 
We quantize pressure data over fixed t-second windows for this process, generating consistent token sequences aligned with F frames per window. To ensure uniform input lengths for batch processing, variable-length pressure sequences are first segmented into t-second windows, with any final segment containing fewer than F frames padded with null values. This method ensures consistent input lengths across all segments, enabling efficient quantization and batch processing.

The pressure sequence, \( p_{1:N} \in \mathbb{R}^{N \times M} \), where \( N \) is the total number of frames and \( M \) is the dimensionality of the spatial pressure representation, is first segmented into fixed t-second intervals of $F$ frames, i.e., \( p^{(i)}_{1:F} \). Each segment is processed individually.

The pressure segment is first encoded into a lower-dimensional latent space using a convolutional encoder \( E \), resulting in a latent vector \( \tilde{p}_{1:n} \in \mathbb{R}^{n \times m} \), where \( n \) is the downsampled number of latent vectors, and \( m \) is their dimensionality. We employ a residual quantization strategy over multiple quantization layers to reduce the information loss inherent in traditional VQ-VAE models.

The first residual $r_0 = \tilde{p}$ is quantized using a predefined codebook $C_0$, iteratively refining the representation across multiple layers. At each layer $v$, the quantization follows:

\begin{equation}
b_v = Q(r_v), \quad r_{v+1} = r_v - b_v
\end{equation}

where $Q(r_v)$ denotes the quantized residual. The cumulative quantized representation is:

\begin{equation}
\tilde{p}{quantized} = \sum{v=0}^{V} b_v
\end{equation}

A unique end token $V_x$ is introduced to delineate sequence boundaries for variable-length sequences. The loss function balances reconstruction fidelity with quantization accuracy using the following:

\begin{equation}
L_{RQVAE} = |p - \hat{p}|1 + \sum{v=0}^{V} \beta_v |r_v - \text{sg}[b_v]|_2^2
\end{equation}

These refinements streamline the methodology while retaining the essential principles of RQ-VAE.

\paragraph{Quantization Dropout}
We also employed quantization dropout to encourage each quantization layer to capture as much helpful information as possible. Ideally, the early quantization layers should restore the most important details in the pressure dynamics. 
In contrast, the later layers refine the missing finer details. To exploit the total capacity of each quantizer, we randomly turn off the last \( 0 \) to \( V \) layers with a probability \( q \in [0,1] \) during training. 
This strategy forces the earlier quantization layers to focus on restoring the critical information.

\subsection{Text2Pressure Generation}
\begin{figure}[!t]
\centering
\includegraphics[width=1\textwidth]{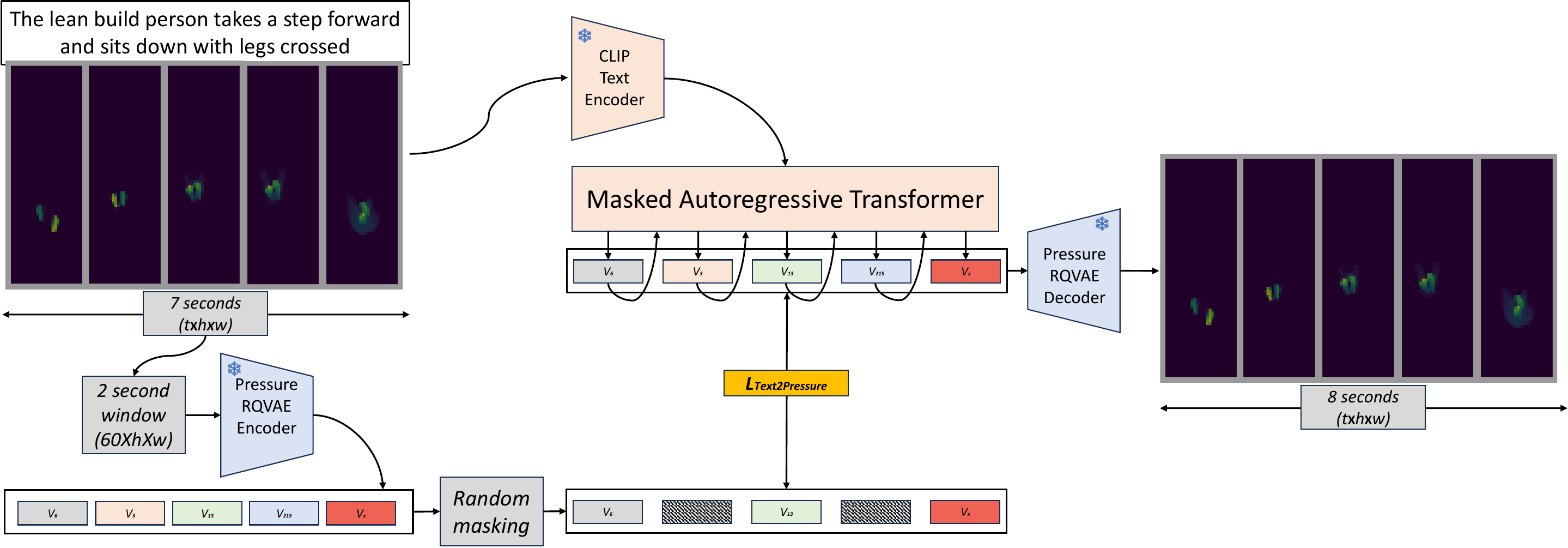} 
\caption{
Text2Pressure uses a frozen CLIP text encoder to convert text into vector embeddings, which are then given as input to a trainable autoregressive transformer to generate \textit{T} tokens up to  \( V_x \). The generated tokens are fed into a pre-trained PressureRQVAE decoder to produce the pressure dynamics.} 
\label{Text2Pressure}
\end{figure}

With a learned PressureRQVAE, a pressure dynamics sequence \( P = [p_1, p_2, ..., p_N] \) can be mapped to a sequence of indices \( S = [s_1, s_2, ..., s_{N/l}, V_x] \), where \( s_i \in \{1, 2, ..., |C|\} \) represents the indices from the learned pressure codebook. Here, a special end token \( V_x \) is added to indicate the stop of the pressure sequence. 
Unlike previous approaches that utilize an additional module to predict sequence length, \( V_x \) allows for a straightforward termination signal in the sequence generation process.

By projecting these indices \( S \) back to their corresponding pressure codebook entries, we obtain the quantized embeddings \( \hat{Z} = [\hat{z}_1, \hat{z}_2, ..., \hat{z}_{N/l}] \), where each \( \hat{z}_i = C_{s_i} \) is the codebook entry corresponding to the index \( s_i \). 
These quantized embeddings can then be decoded through the decoder to reconstruct the pressure sequence \( P_{\text{recon}} \).

The Text2Pressure task aims to generate a pressure sequence conditioned on a given text description, as visualized in \cref{Text2Pressure}. The problem is formulated as an autoregressive next-index prediction task. Given a text condition \( c \) and the previous indices \( S_{<i} \), the model predicts the distribution of the following possible index \( p(s_i | c, S_{<i}) \). This task is addressed using a transformer model, which leverages its causal self-attention mechanism to ensure that the prediction of the current index only depends on the preceding indices.

\paragraph{Loss Function}

The likelihood of the full sequence of indices \( S \) given the text condition \( c \) is represented as:

\begin{equation}
p(S | c) = \prod_{i=1}^{|S|} p(s_i | c, S_{<i})
\end{equation}

During training, we aim to maximize the log-likelihood of the data distribution:

\begin{equation}
L_{\text{Text2Pressure}} = \mathbb{E}_{S \sim p(S)} \left[ - \log p(S | c) \right]
\end{equation}

We leverage the pre-trained CLIP \cite{radford2021learning} model to extract the text embedding \( c \), which has demonstrated effectiveness in cross-modal tasks involving text and other modalities.

The transformer model uses causal self-attention to autoregressively predict the following index, ensuring each token attends only to previous ones via a causal mask. During inference, generation starts from the text embedding and continues until $ V_x $ is predicted.

\paragraph{Random masking}

A common issue arises from the discrepancy between training and inference. During the training procedure, the model uses the ground-truth indices \( S \) as the condition to predict the following index. However, the predicted indices might contain errors during inference, which can propagate through the sequence generation. 
We adopt a data augmentation strategy to mitigate this issue by randomly replacing a percentage \( \tau \times 100\% \) of the ground-truth indices with random indices during training. This strategy helps the model become robust to potential prediction errors during inference, improving overall generation performance.

\subsection{Pressure2Text Inference}

\begin{figure}[!t]
\centering
\includegraphics[width=0.75\textwidth]{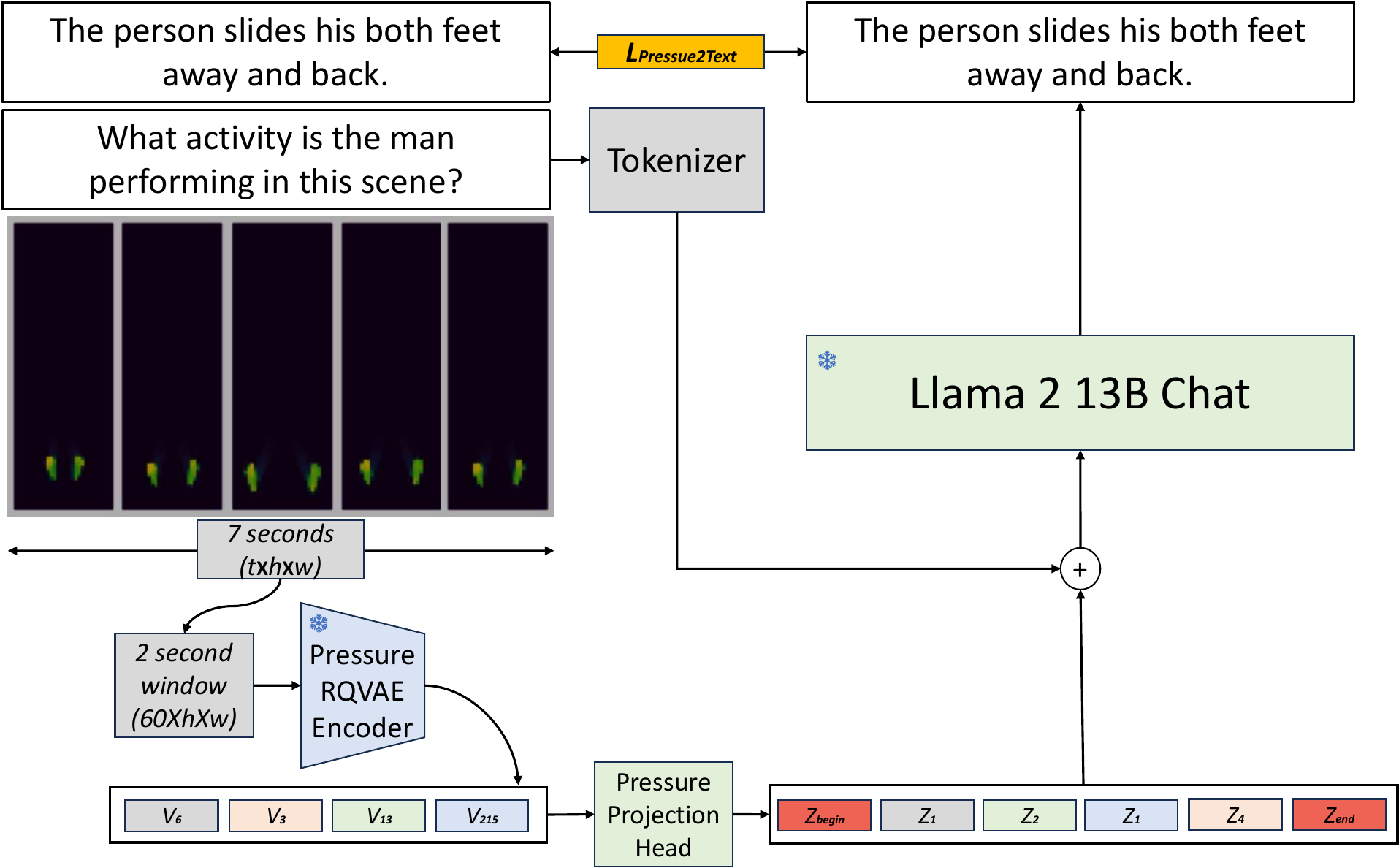} 
\caption{Pressure2Text utilizes a pre-trained PressureRQVAE encoder to transform pressure dynamics into NLP-like tokens. These tokens pass through a trainable projection head, aligning with text tokens. Both the aligned pressure tokens and text tokens are then fed into a frozen LLaMA 2 13B Chat model.} 
\label{PressureRQVAE2}
\end{figure}

With the pre-trained PressureRQVAE, a sequence of discrete pressure tokens \( S = [s_1, s_2, ..., s_{N/l}, V_x] \) can be generated from a pressure dynamics sequence \( P = [p_1, p_2, ..., p_N] \), where \( s_i \in \{1, 2, ..., |C|\} \) represents the indices from the learned pressure codebook. 

These pressure tokens are then embedded into a learned representation \( \hat{Z} = [Z_{\text{begin}}, \hat{z}_1, \hat{z}_2, ..., \hat{z}_{N/l}, Z_{\text{end}}] \) through a projection head, where each \( \hat{z}_i = C_{s_i} \) corresponds to the codebook entry associated with the index \( s_i \). 
At the same time, the text query is tokenized into text tokens. The Pressure projection head aligns the pressure embeddings with the text tokens, creating compatible representations for further processing.
The token \( Z_{\text{begin}} \) is added at the start of the pressure sequence, marking the shift from the text input to the pressure dynamics. 
Conversely, \( Z_{\text{end}} \) is appended at the end of the pressure sequence, signaling the return to the textual representation.
All tokens are then given as input to a frozen LLaMA 13B Chat.

The Pressure2Text task is framed as a next-token prediction problem, similar to how LLMs work. 
Given the aligned pressure embeddings \( \hat{Z} \) and the text tokens \( T_{<i} \), the model predicts the distribution of the following text token \( p(T_i | \hat{Z}, T_{<i}) \) as visualized in \cref{PressureRQVAE2}. 
This autoregressive nature enables the model to generate coherent and contextually relevant text directly informed by the pressure dynamics.

\paragraph{Loss Function}

The loss function for the Pressure2Text task is cross-entropy loss. The likelihood of the generated text sequence \( T \) conditioned on the pressure embeddings \( \hat{Z} \) is represented as:

\begin{equation}
p(T | \hat{Z}) = \prod_{i=1}^{|T|} p(t_i | \hat{Z}, T_{<i})
\end{equation}

During training, we aim to minimize the negative log-likelihood of the generated text:

\begin{equation}
L_{\text{Pressure2Text}} = - \mathbb{E}_{T \sim p(T)} \left[ \log p(T | \hat{Z}) \right]
\end{equation}

The embeddings \( \hat{Z} \) are extracted from the VQ-VAE, while the text tokens are obtained from the tokenized natural language descriptions.

\section{Evaluation}

\begin{figure}[!t]
\centering
\includegraphics[width=1\textwidth]{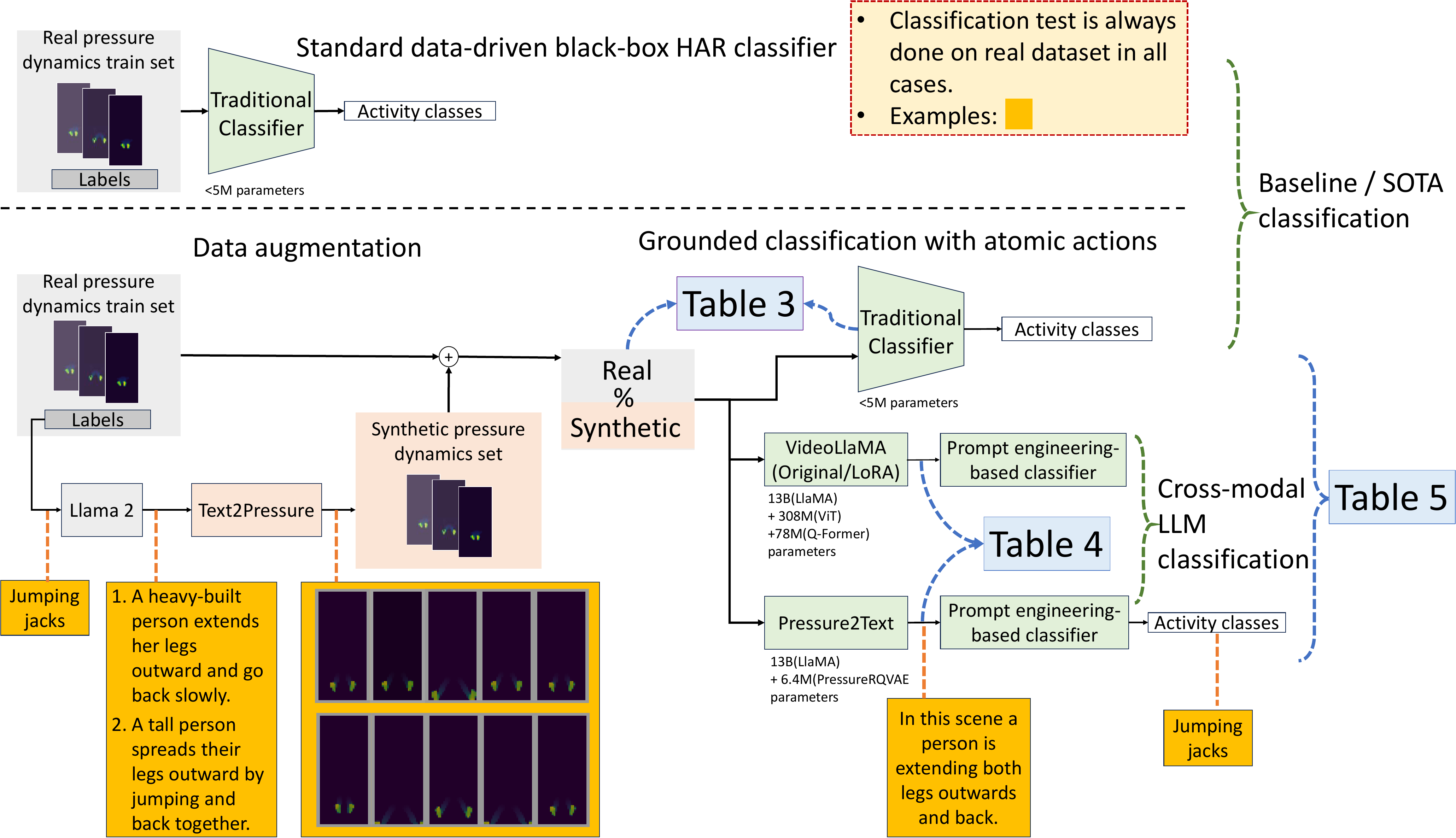} 
\caption{We showcase how both components of TxP can be utilized to create a better HAR system where LLaMA 2, along with Text2Pressure, is used to create a synthetic dataset with variations from the original activity labels while Pressure2Text along with prompt engineering, is utilized to generate an activity label from a dynamic pressure map sequence. We evaluated improvement on HAR by using our data augmentation method (Text2Pressure), grounded classification with atomic action (Pressure2Text), and both together as compared to the contemporary HAR system that uses a traditional classifier (our baseline and other SOTA model on different dataset) to predict the activity label from a fixed window of pressure sensor sequence.} 
\label{overview_eval}
\end{figure}

\label{sec:4}
\subsection{Training Details}
The training of the TxP model involves organizing the dataset into batches for efficient processing, with a batch size of
32 for PressureRQVAE, Pressure2Text, and Text2Pressure tasks to optimize training performance. The PressLang dataset is split into training, validation, and test sets with proportions of 
80 \%, 5 \%, and 15 \%, respectively, ensuring that both the validation and test sets contain labels/activity descriptions unseen by the training set. All the instances are trained for up to a total of \( 500 \) epochs, allowing extensive exposure to the data while monitoring for overfitting. A learning rate of \( 1 \times 10^{-4} \) is utilized, with the ReduceLROnPlateau scheduler employed to adjust the learning rate, where the learning rate is reduced by half if the validation loss does not improve for \( 15 \) consecutive epochs. Early stopping is implemented to prevent overfitting by monitoring the validation loss and halting training if no improvement is observed for \( 30 \) consecutive epochs. The training is conducted on an AdaA6000 Lovelace GPU, utilizing Hugging Face Transformers for the pre-trained CLIP and LLaMA2 13B weights and PyTorch for implementation, facilitating the efficient training and implementation of the model.

\subsection{Evaluation Datasets}
 We evaluated the performance of both Text2Pressure and Pressure2Text across five diverse datasets:

\begin{itemize}
    \item PresSim dataset \cite{ray2023pressim}: This dataset contains 21 complex yoga poses done by 9 participants and captured by SensingTex pressure mattress with \textit{80$\times$28} sensor array characterized by subtle posture and body alignment differences. It requires fine-grained discrimination of pressure patterns to differentiate between similar poses.

    \item TMD dataset \cite{ray2024text}: It comprises eight daily routine activities (e.g., walking, sitting, standing) that vary significantly in pressure distribution across body regions. It is also captured by SensingTex pressure mattress with \textit{80$\times$28} sensor array for 10 participants.
    
    \item PmatData \cite{pouyan2017pressure}: This dataset focuses on 11 different in-bed sleeping postures. These postures involve subtle but distinct shifts, making them challenging to recognize. It was collected using Vista Medical FSA SoftFlex 2048 with size \textit{32$\times$64}. 13 participants were used to capture this dataset.

    \item MeX \cite{wijekoon2019mex}: Featuring seven physiotherapy exercises and 30 participants, the MeX dataset tests the model's capacity to recognize therapeutic and rehabilitative movements. It was captured using a SensingTex mattress with \textit{32*16} sensor array.

    \item PID4TC \cite{o2024ai}: A pressure insole dataset containing five activities: standing, walking, pick and place, assembly, and manual handling. The sensors are located on the heel, midfoot, and forefoot of the insoles, occupying 30\%, 30\%, and 40\% of the insole length, respectively, with a sampling rate of 100Hz.
\end{itemize}

\subsection{Evaluation Metrics}
We evaluate three aspects of the TxP framework: quality of dynamic pressure maps, quality of the text, and classification performance specific to the dataset using the different matrices.

For assessing the spatial quality of dynamic pressure maps reconstructed back from the Pressure codebook using PressureRQVAE and the quality of generated pressure dynamics by Text2Pressure, we use the following metrics:  
\begin{itemize}
    \item Fréchet Inception Distance (FID): It assesses the quality of generated pressure dynamics by measuring the distance between real and generated pressure token distributions:
    \begin{equation}
    FID(\mu_r, \Sigma_r, \mu_g, \Sigma_g) = \|\mu_r - \mu_g\|^2 + \text{Tr}(\Sigma_r + \Sigma_g - 2(\Sigma_r \Sigma_g)^{1/2}),
    \end{equation}
    where \( \mu \) and \( \Sigma \) represent the mean and covariance of the respective distributions. 
    
    \item \( R^2 \) Score: This measures the proportion of variance explained by the model's predictions, where higher scores show better accuracy. 
    
    \item Binarized \( R^2 \): Adapted from PresSim \cite{ray2023pressim}, this metric evaluates binary thresholded pressure maps highlighting spatial fidelity.    
\end{itemize}

For assessing the quality of encoding and retrieval of tokens generated for the dynamic pressure maps using PressureRQVAE and Text2Pressure, we use the following metric:  

\begin{itemize}
    \item Relevant Precision (R-Precision): It is a ranking-based metric primarily used in tasks that involve retrieval, such as similarity matching.
    This metric evaluates how well-generated tokens align with expected outcomes at different levels. 
    Higher values indicate coherent pressure token sequences.

\end{itemize}

\begin{table}[!t]
    \footnotesize
    \centering
    \caption{Comparison of VQVAE from TMD with PressureRQVAE (with and without quantization dropout) across different codebook sizes, window sizes, and with or without quantization dropout on the PressLang test set.}
    \begin{tabular}{|c|c|c|c|c|c|c|}
        \hline
        \textbf{Model} & \textbf{Input: Window Size (30 Hz)} & \textbf{Bottleneck: Codebook} & \textbf{FID} $\downarrow$ & \textbf{R Precision 1} $\uparrow$ & \textbf{R Precision 5} $\uparrow$ \\ 
        \hline
        \multirow{12}{*}{TMD:VQVAE} & \multirow{4}{*}{30$\times80\times$28} & 128 & $1.544 \pm 0.014$ & $0.225 \pm 0.012$ & $0.352 \pm 0.010$ \\ 
        & & 256 & $1.108 \pm 0.013$ & $0.289 \pm 0.011$ & $0.406 \pm 0.009$ \\ 
        & & 512 & $0.719 \pm 0.010$ & $0.338 \pm 0.010$ & $0.470 \pm 0.012$ \\ 
        & & 1024 & $0.598 \pm 0.009$ & $0.416 \pm 0.013$ & $0.563 \pm 0.015$ \\ 
        \cline{2-6}
        & \multirow{4}{*}{60$\times80\times$28} & 128 & $1.286 \pm 0.012$ & $0.288 \pm 0.015$ & $0.403 \pm 0.009$ \\ 
        
        & & 256 & $0.717 \pm 0.010$ & $0.354 \pm 0.012$ & $0.481 \pm 0.011$ \\ 
        
        & & 512 & $0.408 \pm 0.008$ & $0.426 \pm 0.010$ & $0.545 \pm 0.013$ \\ 
        
        & & 1024 & $0.336 \pm 0.013$ & $0.502 \pm 0.011$ & $0.644 \pm 0.014$ \\ 
        \cline{2-6}
        & \multirow{4}{*}{120$\times80\times$28} & 128 & $1.582 \pm 0.015$ & $0.234 \pm 0.013$ & $0.367 \pm 0.011$ \\ 
        
        & & 256 & $1.218 \pm 0.013$ & $0.292 \pm 0.012$ & $0.417 \pm 0.010$ \\ 
        
        & & 512 & $0.861 \pm 0.011$ & $0.352 \pm 0.011$ & $0.484 \pm 0.012$ \\ 
        
        & & 1024 & $0.656 \pm 0.010$ & $0.429 \pm 0.014$ & $0.572 \pm 0.016$ \\ 
        \hline
        \multirow{12}{*}{PressureRQVAE} & \multirow{4}{*}{30$\times80\times$28} & 128 & $1.513 \pm 0.014$ & $0.252 \pm 0.012$ & $0.366 \pm 0.011$ \\ 
        
        & & 256 & $1.007 \pm 0.013$ & $0.313 \pm 0.011$ & $0.421 \pm 0.010$ \\ 
        
        & & 512 & $0.680 \pm 0.010$ & $0.366 \pm 0.011$ & $0.492 \pm 0.012$ \\ 
        
        & & 1024 & $0.578 \pm 0.009$ & $0.441 \pm 0.013$ & $0.608 \pm 0.015$ \\ 
        \cline{2-6}
        & \multirow{4}{*}{60$\times80\times$28} & 128 & $1.251 \pm 0.012$ & $0.303 \pm 0.015$ & $0.422 \pm 0.010$ \\ 
        
        & & 256 & $0.681 \pm 0.010$ & $0.377 \pm 0.012$ & $0.494 \pm 0.011$ \\ 
        
        & & 512 & $0.381 \pm 0.008$ & $0.442 \pm 0.010$ & $0.577 \pm 0.013$ \\ 
        
        & & 1024 & $0.311 \pm 0.013$ & $0.522 \pm 0.011$ & $0.670 \pm 0.014$ \\ 
        \cline{2-6}
        & \multirow{4}{*}{120$\times80\times$28} & 128 & $1.455 \pm 0.015$ & $0.252 \pm 0.013$ & $0.368 \pm 0.011$ \\ 
        
        & & 256 & $1.097 \pm 0.013$ & $0.329 \pm 0.012$ & $0.432 \pm 0.010$ \\ 
        
        & & 512 & $0.843 \pm 0.011$ & $0.362 \pm 0.011$ & $0.496 \pm 0.012$ \\ 
        
        & & 1024 & $0.649 \pm 0.010$ & $0.443 \pm 0.014$ & $0.588 \pm 0.016$ \\ 
        \hline
        \multirow{8}{*}{PressureRQVAE} & \multirow{4}{*}{30$\times80\times$28} & 128 & $1.052 \pm 0.012$ & $0.363 \pm 0.012$ & $0.482 \pm 0.011$ \\ 
        
        & & 256 & $0.602 \pm 0.011$ & $0.431 \pm 0.010$ & $0.525 \pm 0.012$ \\ 
        
        & & 512 & $0.253 \pm 0.009$ & $0.498 \pm 0.011$ & $0.602 \pm 0.012$ \\ 
        
        & & 1024 & $0.151 \pm 0.008$ & $0.571 \pm 0.013$ & $0.699 \pm 0.015$ \\ 
        \cline{2-6}
        & \multirow{4}{*}{60$\times80\times$28} & 128 & $0.952 \pm 0.010$ & $0.414 \pm 0.014$ & $0.517 \pm 0.010$ \\ 
        
        & & 256 & $0.402 \pm 0.009$ & $0.495 \pm 0.011$ & $0.591 \pm 0.011$ \\ 
        
        +& & 512 & $0.152 \pm 0.008$ & $0.577 \pm 0.010$ & $0.681 \pm 0.013$ \\ 
        
        & & 1024 & \textbf{0.095 $\pm$ 0.012} & \textbf{0.656 $\pm$ 0.012} & \textbf{0.791 $\pm$ 0.014} \\ 
        \cline{2-6}
        quantization& \multirow{4}{*}{120$\times80\times$28} & 128 & $1.003 \pm 0.013$ & $0.319 \pm 0.013$ & $0.427 \pm 0.011$ \\ 
        
        dropout& & 256 & $0.552 \pm 0.011$ & $0.390 \pm 0.012$ & $0.517 \pm 0.010$ \\ 
        
        & & 512 & $0.356 \pm 0.010$ & $0.468 \pm 0.011$ & $0.587 \pm 0.012$ \\ 
        
        & & 1024 & $0.253 \pm 0.009$ & $0.541 \pm 0.014$ & $0.677 \pm 0.016$ \\ 
        \hline
        \end{tabular}
        \label{tab:results_comparison}
    \end{table}
    
For assessing the quality of natural language inferred from the pressure dynamics using Pressure2Text, we use the following:

\begin{itemize}
    \item Cosine similarity: It assesses semantic alignment between the generated and ground truth text. 
    While helpful in measuring semantic alignment between generated and ground truth text, cosine similarity has limitations in attribute estimation, particularly when dealing with sensitive or overly specific attributes. LLMs are primarily trained on conversational and descriptive text rather than concise, attribute-only phrases. As a result, we continue to use more descriptive outputs despite this limitation.

    To address this and to focus specifically on evaluating whether the correct activity has been identified, which is the central goal of our model, we apply post-generation prompt engineering to extract only the activity type from the generated descriptive sentence. This filtered output allows us to evaluate activity classification performance using the macro F1 score, offering a more precise and task-aligned metric for assessing model accuracy.

\end{itemize}

Finally, for assessing the activity classification performance specific to different datasets using either Text2Pressure, Pressure2Text, or both, we utilize the following:
\begin{itemize}
    \item Macro F1 Score: It is commonly used for estimating HAR model performance while considering the class balance.
\end{itemize}

\begin{figure}[!t]
\centering
\includegraphics[width=1\textwidth]{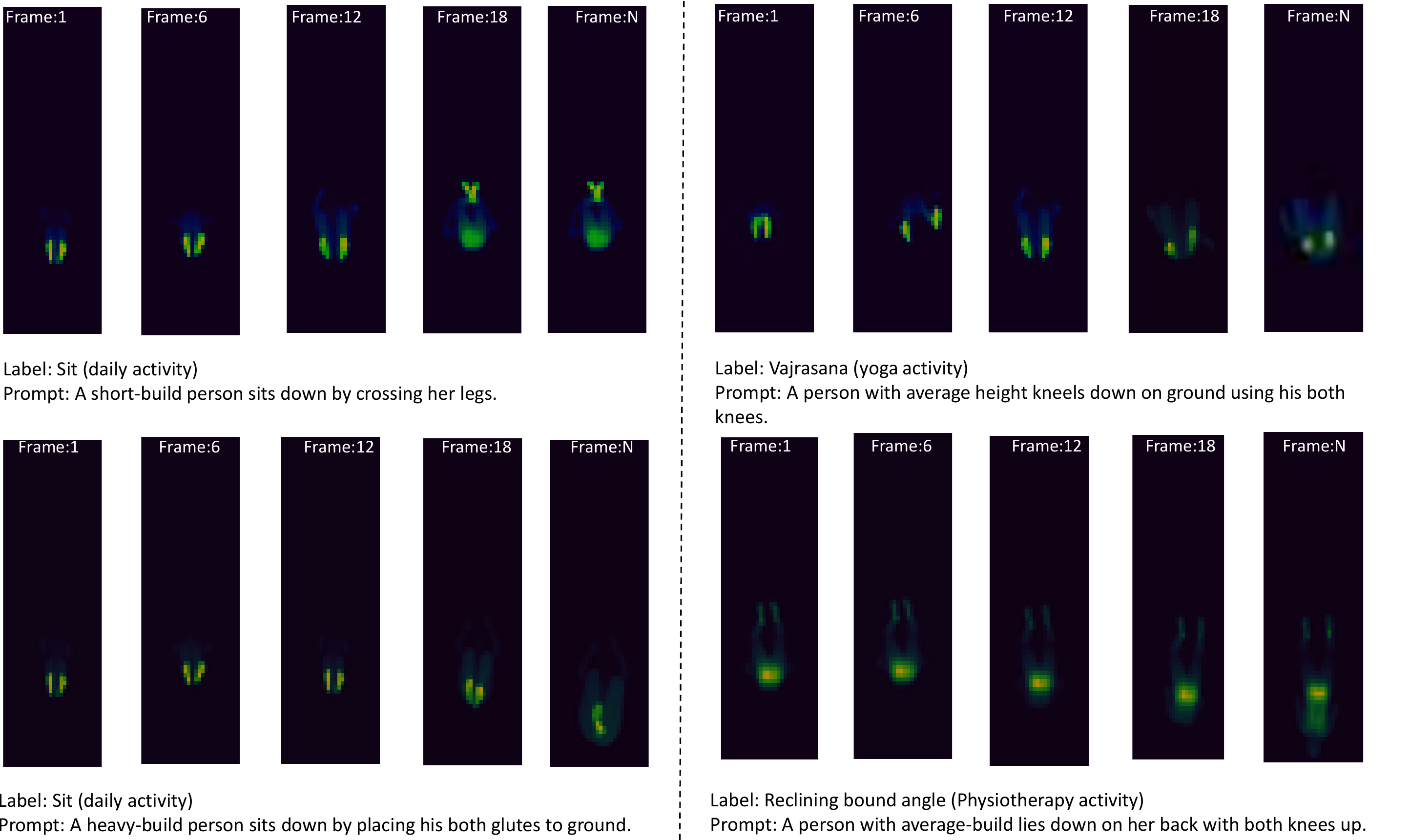} 
\caption{Qualitative results showcasing generating the same activity with different variations (sitting) as well as different activities (yoga pose and physiotherapy exercise) using Text2Pressure.} 
\label{qual_eval}
\end{figure}

\subsection{PressureRQVAE Evaluation and Results}
\paragraph{Quantitative assessment} 

To evaluate the PressureRQVAE model, we compared it with the Vector Quantized Variational Autoencoder from TMD \cite{ray2024text} (TMD: VQVAE), as shown in \cref{tab:results_comparison}, focusing on codebook sizes and window lengths.
Using codebook sizes of 128, 256, 512, and 1024, we found that larger sizes improved the capture of fine-grained pressure details. Specifically, a codebook of 1024 yielded consistently lower FID scores, indicating enhanced quality in reconstructing dynamics close to the original data.
For window sizes of \(1\), \(2\), and \(4\) seconds, the \(2\)-second window proved optimal, balancing sufficient temporal context and computational efficiency. While the \(1\)-second window lost temporal context, the \(4\)-second window introduced complexity, affecting quantization accuracy.
PressureRQVAE outperformed TMD: VQVAE, especially with more extensive dictionaries and the \(2\)-second window size. Quantization dropout further boosted results, underscoring its role in robustly quantizing and reconstructing continuous pressure dynamics.
It also validates that even with this kind of discrete quantization, we can still reconstruct the original signal with very high accuracy. Hence, the embedding has enough information for other downstream tasks.

\subsection{Text2Pressure Evaluation and Results}

\paragraph{Qualitative assessment}
We showcased several examples generated by the Text2Pressure model in \cref{qual_eval} to provide a qualitative assessment of its performance.
The model can generate different variations of the same activity, such as sitting, and a different range of activities, from yoga poses to physiotherapy exercises to daily activities, as shown in the figure.
These examples illustrate the model's ability to generate realistic and contextually appropriate pressure patterns based on varied textual descriptions. 
However, we also want to highlight that the generated data are imperfect and exhibit some visual flaws due to the PresSim architecture used to simulate synthetic pressure data, which relies on depth image simulation. 
Enhancing these visual qualities is beyond the scope of this work. Nevertheless, despite these imperfections, the data generated by Text2Pressure contributes to improved HAR scores, as discussed later in the evaluation section. 

\paragraph{Quantitative assessment}

\begin{table}[!t]
\footnotesize
    \centering
    \caption{Comparison of Generated Pressure Data Using Text2Pressure with and without random masking, TMD \cite{ray2024text}, and PressureTransferNet \cite{ray2023pressuretransfernet}.}
    \begin{tabular}{|c|c|c|c|c|c|}
        \hline
        \textbf{Model} & \textbf{FID} $\downarrow$ & \textbf{R Precision 1} $\uparrow$ & \textbf{R Precision 5} $\uparrow$ & \textbf{R$^2$} $\uparrow$ & \textbf{Binarized R$^2$} $\uparrow$ \\ 
        \hline
        Baseline & $5.783 \pm 0.045$ & $0.105 \pm 0.010$ & $0.180 \pm 0.008$ & $0.150 \pm 0.023$ & $0.200 \pm 0.015$ \\
        \hline
        PressureTransferNet \cite{ray2023pressuretransfernet}& - & - & - & $0.614 \pm 0.027$ & $0.691 \pm 0.018$ \\ 
        \hline
        TMD \cite{ray2024text} & $1.830 \pm 0.021 $ & $0.218 \pm 0.016 $ & $0.344 \pm 0.009$ & $0.722 \pm 0.031$ & $0.892 \pm 0.025$ \\ 
        \hline
        Text2Pressure (Proposed)& $1.105 \pm 0.018$ & $0.305 \pm 0.012$ & $0.450 \pm 0.010$ & $0.739 \pm 0.030$ & $0.898 \pm 0.022$ \\ 
        \hline
        Text2Pressure (Proposed) &  &  &  &  &  \\ 
        +& \textbf{1.003 $\pm$ 0.017} & \textbf{0.321 $\pm$ 0.013} & \textbf{0.465 $\pm$ 0.011} & \textbf{0.774 $\pm$ 0.028} & \textbf{0.901 $\pm$ 0.020} \\ 
        Random masking  &  &  &  &  &  \\ 
        \hline
    \end{tabular}
    \label{tab:quality_comparison}
\end{table}

As shown in \cref{tab:quality_comparison}, Text2Pressure outperformed existing pressure generation models, including TMD \cite{ray2024text}, PressureTransferNet \cite{ray2023pressuretransfernet}, and a baseline model identical to Text2Pressure but using unquantized raw pressure dynamics.
Text2Pressure achieved lower FID scores, indicating higher fidelity in generated dynamics, and consistently outperformed in R Precision at 1 and 5, demonstrating superior relevance in retrieved sequences. 
Additionally, its higher $ R2$ and binarized $ R2$ scores highlighted its ability to generate precise and well-distributed pressure values across individual points and entire pressure maps. 
These results confirm Text2Pressure’s effectiveness in creating complex pressure patterns with greater accuracy and contextual coherence than other models.
We also showcased the importance of random masking for training an autoregressive model like Text2Pressure through the evaluation.

\paragraph{Evaluation on Downstream Data Augmentation}

\begin{table}[!t]
\footnotesize
    \centering
    \caption{Comparison of Macro F1 across different datasets and data compositions using different synthetic pressure data.}
    \begin{tabular}{|c|c|c|c|c|c|}
        \hline
        \textbf{Data generator} & \textbf{Composition} & \multicolumn{4}{c|}{\textbf{Macro F1} $\uparrow$}  \\ 
        \hline
        &Real/Synthetic & \textbf{PresSim} \cite{ray2023pressim}& \textbf{TMD dataset} \cite{ray2024text}& \textbf{PmatData} \cite{pouyan2017pressure}& \textbf{MeX} \cite{wijekoon2019mex}\\ 
        \hline
        No synthetic data & 100\% / 0\% & $0.879 \pm 0.014$ & $0.742 \pm 0.038$ & $0.765 \pm 0.026 $ & $0.661 \pm 0.022$ \\
        \hline
        TMD &  & $0.720 \pm 0.018$ & $0.740 \pm 0.020$ & $0.610 \pm 0.019$ & $0.550 \pm 0.020$ \\
        PressureTransferNet & 90\% / 10\% & $0.845 \pm 0.018$ & $0.750 \pm 0.023$ & $0.748 \pm 0.021$ & $0.655 \pm 0.018$ \\
        Text2Pressure &  & $0.860 \pm 0.020 $ & $0.755 \pm 0.025 $ & $0.710 \pm 0.015 $ & $0.630 \pm 0.020$ \\
        \hline
        TMD &  & $0.796 \pm 0.017$ & $0.756 \pm 0.024$ & $0.644 \pm 0.012 $ & $0.590 \pm 0.024$ \\
        PressureTransferNet & 75\% / 25\% & $0.868 \pm 0.016 $ & $0.763 \pm 0.031$ & $0.773 \pm 0.023 $ & $0.651 \pm 0.017$ \\
        Text2Pressure &  & $0.873 \pm 0.024 $ & $0.788 \pm 0.017$ & $0.638 \pm 0.016 $ & $0.665 \pm 0.031$ \\
        \hline
        TMD &  & $0.780 \pm 0.019$ & $0.728 \pm 0.022$ & $0.590 \pm 0.017$ & $0.540 \pm 0.022$ \\
        PressureTransferNet & 60\% / 40\% & $0.880 \pm 0.018$ & $0.785 \pm 0.021$ & $0.765 \pm 0.019$ & $0.660 \pm 0.020$ \\
        Text2Pressure &  & $0.890 \pm 0.022 $ & $0.800 \pm 0.022$ & $0.580 \pm 0.025 $ & $0.690 \pm 0.023$ \\
        \hline
        TMD &  & $0.852 \pm 0.022 $ & $0.711 \pm 0.027$ & $0.542 \pm 0.025 $ & $0.638 \pm 0.025$ \\
        PressureTransferNet & 50\% / 50\% & $0.911 \pm 0.015 $ & $0.801 \pm 0.027$ & \textbf{0.781 $\pm$ 0.013} & $0.702 \pm 0.019$ \\
        Text2Pressure &  & \textbf{0.912 $\pm$ 0.024}  & \textbf{0.858 $\pm$ 0.019} & $0.544 \pm 0.031 $ & \textbf{0.719 $\pm$ 0.027} \\
        \hline
        TMD &  & $0.780 \pm 0.021$ & $0.700 \pm 0.025$ & $0.500 \pm 0.022$ & $0.580 \pm 0.023$ \\
        PressureTransferNet & 40\% / 60\% & $0.865 \pm 0.020$ & $0.770 \pm 0.023$ & $0.720 \pm 0.022$ & $0.645 \pm 0.021$ \\
        Text2Pressure &  & $0.835 \pm 0.025 $ & $0.790 \pm 0.018$ & $0.470 \pm 0.015 $ & $0.675 \pm 0.020$ \\
        \hline
        TMD &  & $0.698 \pm 0.033 $ & $0.698 \pm 0.022 $ &  $0.434 \pm 0.013$  & $0.552 \pm 0.022$ \\
        PressureTransferNet & 25\% / 75\% & $0.712 \pm 0.018 $ & $0.706 \pm 0.015 $ & $0.761 \pm 0.024 $ & $0.596 \pm 0.018$ \\
        Text2Pressure &  & $0.707 \pm 0.027 $ & $0.743 \pm 0.018$ & $0.428 \pm 0.014 $ & $0.641 \pm 0.015$  \\
        \hline
        TMD &  & $0.650 \pm 0.025$ & $0.630 \pm 0.022$ & $0.400 \pm 0.018$ & $0.510 \pm 0.020$ \\
        PressureTransferNet & 10\% / 90\% & $0.720 \pm 0.020$ & $0.700 \pm 0.018$ & $0.690 \pm 0.017$ & $0.570 \pm 0.022$ \\
        Text2Pressure &  & $0.600 \pm 0.025 $ & $0.670 \pm 0.020 $ & $0.340 \pm 0.018 $ & $0.610 \pm 0.025$ \\
        \hline
        TMD &  & $0.416 \pm 0.024 $ & $0.502 \pm 0.025$ & $0.216 \pm 0.036 $ & $0.450 \pm 0.025$ \\
        PressureTransferNet & 0\% / 100\% & $0.803 \pm 0.002$ & $0.561 \pm 0.018$ & $0.656 \pm 0.017 $ & $0.531 \pm 0.018$ \\
        Text2Pressure &  & $0.482 \pm 0.021 $ & $0.642 \pm 0.031$ & $0.244 \pm 0.028$ & $0.575 \pm 0.031$ \\
        \hline
    \end{tabular}

    \label{tab:har_comparison_f1}
\end{table}

To evaluate Text2Pressure for HAR, we generated synthetic data using diverse descriptions of identical activities, facilitating the expansion of activity data through varied but realistic linguistic expressions. These descriptions were created using LLaMA 2, adding slight variations that mirror how real activities may be described differently across contexts.

For example, taking the activity of \textit{"jumping jacks"}, we used descriptions like: \textit{"A heavy-built person extends her legs outward and goes back slowly."} and \textit{"A tall person spreads their legs outward by jumping and brings them back together quickly."}
The variations in description impact the generated synthetic pressure data in several ways. Other than the noticeable body-associated changes for the "slow" description, Text2Pressure would allocate a more significant number of tokens to each atomic motion, representing an extended duration of each movement, which simulates the slower pace of the action. 
Conversely, for the "fast" description, Text2Pressure would use fewer tokens per atomic motion, condensing each movement phase. This approach effectively adjusts the temporal granularity and speed of the activity, capturing subtle variations in movement speed and rhythm through token allocation.
These variations, therefore, allow the synthetic pressure data to capture different activity paces, helping the model generalize better to real-world scenarios where the same action may be performed with different dynamics, such as a person moving faster or slower based on physical characteristics or context.

We used the 3D CNN model from PressureTransferNet \cite{ray2023pressuretransfernet} as a baseline classifier to evaluate our model on the four public datasets as given in \cref{tab:har_comparison_f1}.
We also conducted an ablation study to assess how varying proportions of synthetic and real data affect HAR model performance. 
Training data included different ratios: 50 \% real and 50 \% synthetic data, 10 \% real and 90 \% synthetic, 25 \% real and 75 \% synthetic, 40 \% real and 60 \% synthetic, 75 \% real and 25 \% synthetic, 60 \% real and 40 \% synthetic, 90 \% real and 10 \% synthetic, and a purely synthetic set.
The 50 \% real and 50 \% synthetic configurations achieved the highest performance across datasets, suggesting that balanced synthetic data enhances model understanding of activity variations. 
The HAR score is increased when the percentage of real data is higher or similar to synthetic data. 
However, a higher level of synthetic data led to accuracy drops, likely due to overfitting, and training on 100 \% synthetic data resulted in even lower performance, likely because of the unfamiliarity of the model with real data. 
This study emphasizes the need for a balanced real-to-synthetic ratio for optimal model robustness in real-world scenarios.

In comparison with other synthetic data generation models like TDM \cite{ray2024text}, which synthesizes pressure data from text descriptions, and PressureTransferNet \cite{ray2023pressuretransfernet}, which augments real pressure data for varying body types, the HAR model trained with Text2Pressure data achieved superior performance. It exceeded real-only data on PreSim, TMD dataset, and MeX by 3.3 \%, 11.6 \%, and 5.8 \%, respectively. 
Compared to the state-of-the-art augmentation method PressureTransferNet, it showed gains of 0.1 \%, 5.7 \%, and 1.7 \%, respectively. 
This can likely be attributed to the absence of sleeping activities in the PressLang dataset, a limitation from which PressureTransferNet, being based on real pressure data augmentations, does not suffer, thus achieving better results for this specific dataset.

We also want to point out that although PressureTransferNet achieved a higher score than TxP, it has a pronounced limitation. Unlike TxP, it requires at least some real data from the particular sensor array for each target activity to generate more pressure dynamics, making it impractical for most use cases.

\subsection{Pressure2Text Evaluation and Results}

\paragraph{Quantitative Assessment}

\begin{table}[!t]
\footnotesize
    \centering
    \caption{Comparison of Cosine similarity across different datasets with and without using Text2Pressure with VideoLLaMA \cite{zhang2023video}, LoRA VideoLLaMA and Pressure2Text.}
    \begin{tabular}{|c|c|c|c|c|c|}
        \hline
        \multirow{2}{*}{\textbf{Classifier}} & \multirow{2}{*}{\textbf{Training Set}} & \multicolumn{4}{c|}{\textbf{Cosine similarity} $\uparrow$}  \\ 
        \cline{3-6}
        & & \textbf{PresSim \cite{ray2023pressim}} & \textbf{TMD dataset} \cite{ray2024text} & \textbf{PmatData} \cite{pouyan2017pressure}& \textbf{MeX} \cite{wijekoon2019mex}\\
        \hline
        \multirow{2}{*}{VideoLLaMA \cite{zhang2023video}} & Real & $0.137 \pm 0.016$ & $0.184 \pm 0.022$ & $0.201 \pm 0.021 $ & $0.210 \pm 0.014$ \\
        \cline{2-6}
        & Real + Text2Pressure & $0.155 \pm 0.024$ & $0.231 \pm 0.021 $ & $0.206 \pm 0.011 $ & $ 0.288 \pm 0.008$ \\
        \hline
        \multirow{2}{*}{VideoLLaMA (LoRA)} & Real & $0.197 \pm 0.023 $ & $0.208 \pm 0.025 $ & $0.254 \pm 0.013 $ & $0.246 \pm 0.016 $ \\
        \cline{2-6}
        & Real + Text2Pressure & $0.203 \pm 0.008$ & $0.296 \pm 0.017 $ & $0.248 \pm 0.019$ & $0.312 \pm 0.014 $ \\
        \hline
        \multirow{2}{*}{Pressure2Text} & Real & $0.445 \pm 0.026 $ & $0.588 \pm 0.023 $ & \textbf{0.263 $\pm$ 0.021} & $0.552 \pm 0.015 $ \\
        \cline{2-6}
        & Real + Text2Pressure & \textbf{0.451 $\pm$ 0.014}  & \textbf{0.633 $\pm$ 0.023} & $0.261 \pm 0.016 $ & \textbf{0.571 $\pm$ 0.029} \\
        \hline
    \end{tabular}
    \label{tab:har_comparison_cosine}
\end{table}

We compared Pressure2Text with two alternative cross-modal text generation approaches based on LLM backbones: VideoLLaMA \cite{zhang2023video}, which can generate text from audio/video and text data, and a Low-Rank Adapted (LoRA) version \cite{hu2021lora} of VideoLLaMA. For a fair comparison of the PressLang dataset, we converted the pressure dynamics into an image format compatible with their inputs.

Pressure2Text demonstrated superior results in terms of cosine similarity, as shown in \cref{tab:har_comparison_cosine}, despite using approximately 500 million fewer parameters than its counterparts. However, the results were not perfect—variations in phrasing can lead to semantically equivalent outputs that still score low in cosine similarity due to differences in grammar, word choice, or sentence structure.

In the case of PmatData, as discussed in the Text2Pressure evaluation, the model performed worse when trained on synthetic data than real data. This drop in performance is primarily due to the inaccuracy of synthetic pressure labels for sleeping poses. To more accurately evaluate our model, we implemented a classifier using prompt engineering within LLaMA 2 13B Chat, as described in the next section.

While using cosine similarity for attribute estimation may not reliably capture the correctness of sensitive attributes, we addressed this by computing the F1 score from sentences.
Inspired by findings in works like OVHHIR \cite{10890689}, we observed that training LLMs to generate complete descriptive sentences rather than rigidly structured phrases significantly improved generation quality. 
Due to LLMs' training in natural language, they are more effective at sentence-level generation. 
To extract structured information such as activity labels, we use post-processing with the LLaMA model itself—extracting activities from the generated sentences, converting them into one-hot encoded vectors, and finally computing F1 scores for robust evaluation.

\paragraph{Evaluation on Downstream Classification}
\begin{table}[!t]
\footnotesize
    \centering
    \caption{Comparison of Macro F1 across Different Datasets and Data Compositions for baseline classifier from and prompt engineered classifiers over cross-modal LLM backbones. SOTA results reported in the table are from the original papers.}
    \begin{tabular}{|c|c|c|c|c|c|}
        \hline
                \multirow{2}{*}{\textbf{Classifier}} & \multirow{2}{*}{\textbf{Training Set}} & \multicolumn{4}{c|}{\textbf{Macro F1} $\uparrow$}  \\ 
        \cline{3-6}
        & & \textbf{PresSim \cite{ray2023pressim}} & \textbf{TMD dataset} \cite{ray2024text} & \textbf{PmatData} \cite{pouyan2017pressure}& \textbf{MeX} \cite{wijekoon2019mex}\\
        \hline
        \multirow{1}{*}{SOTA$^*$} & (100\%)Real & $0.879 \pm 0.014$ & $0.737 \pm 0.035$ & \textbf{0.782 $\pm$ 0.018} & $0.719$ \\
        \hline
        \multirow{2}{*}{Baseline} & (100\%)Real & $0.878 \pm 0.025$ & $0.742 \pm 0.038 $ & 0.765 $\pm$ 0.026 & $0.661 \pm 0.022$ \\
        & (90\%)Real + (10\%)Text2Pressure & $0.860 \pm 0.020$ & $0.755 \pm 0.025$ & $0.710 \pm 0.015$ & $0.630 \pm 0.020$ \\
        & (75\%)Real + (25\%)Text2Pressure & 0.873 ± 0.024 & 0.788 ± 0.017 & 0.638 ± 0.016 & 0.665 ± 0.031 \\
        & (60\%)Real + (40\%)Text2Pressure & $0.890 \pm 0.022$ & $0.800 \pm 0.022$ & $0.580 \pm 0.025$ & $0.690 \pm 0.023$ \\
        & (50\%)Real + (50\%)Text2Pressure & \textbf{0.912 $\pm$ 0.024} & $0.858 \pm 0.019 $ & $0.544 \pm 0.031 $ & $0.719 \pm 0.027$ \\
        & (40\%)Real + (60\%)Text2Pressure & $0.835 \pm 0.025$ & $0.790 \pm 0.018$ & $0.470 \pm 0.015$ & $0.675 \pm 0.020$ \\
        & (25\%)Real + (75\%)Text2Pressure  & 0.707 ± 0.027 &  0.743 ± 0.018 & 0.428 ± 0.014 &  0.641 ± 0.015 \\
        & (10\%)Real + (90\%)Text2Pressure & $0.600 \pm 0.025$ & $0.670 \pm 0.020$ & $0.340 \pm 0.018$ & $0.610 \pm 0.025$ \\
        & (100\%)Text2Pressure  & 0.482 ± 0.021 &  0.642 ± 0.031 &  0.244 ± 0.028 & 0.575 ± 0.031 \\
        \hline
        \multirow{2}{*}{VideoLLaMA \cite{zhang2023video}} & (100\%)Real & $0.413 \pm 0.026$ & $0.502 \pm 0.019$ & $0.567 \pm 0.024$ & $0.445 \pm 0.016$ \\
        & (10\%)Real + (90\%)Text2Pressure & $0.360 \pm 0.024$ & $0.470 \pm 0.020$ & $0.420 \pm 0.022$ & $0.410 \pm 0.018$ \\
        & (25\%)Real + (75\%)Text2Pressure & $0.395 \pm 0.022$ & $0.490 \pm 0.018$ & $0.460 \pm 0.021$ & $0.430 \pm 0.017$ \\
        & (40\%)Real + (60\%)Text2Pressure & $0.420 \pm 0.020$ & $0.520 \pm 0.017$ & $0.500 \pm 0.020$ & $0.460 \pm 0.016$ \\
        & (50\%)Real + (50\%)Text2Pressure & $0.449 \pm 0.018$ & $0.545 \pm 0.021$ & $0.483 \pm 0.023$ & $0.493 \pm 0.017$ \\
        & (60\%)Real + (40\%)Text2Pressure & $0.430 \pm 0.019$ & $0.530 \pm 0.020$ & $0.460 \pm 0.021$ & $0.475 \pm 0.016$ \\
        & (75\%)Real + (25\%)Text2Pressure & $0.420 \pm 0.020$ & $0.515 \pm 0.019$ & $0.445 \pm 0.020$ & $0.460 \pm 0.017$ \\
        & (90\%)Real + (10\%)Text2Pressure & $0.415 \pm 0.021$ & $0.505 \pm 0.018$ & $0.430 \pm 0.019$ & $0.450 \pm 0.017$ \\
        & (100\%)Text2Pressure & $0.340 \pm 0.025$ & $0.460 \pm 0.022$ & $0.390 \pm 0.024$ & $0.420 \pm 0.019$ \\
        \hline
        \multirow{2}{*}{VideoLLaMA (LoRA)} & (100\%)Real & $0.502 \pm 0.021$ & $0.531 \pm 0.023$ & $0.515 \pm 0.022$ & $0.488 \pm 0.015$ \\
        & (10\%)Real + (90\%)Text2Pressure & $0.400 \pm 0.024$ & $0.480 \pm 0.020$ & $0.410 \pm 0.021$ & $0.440 \pm 0.018$ \\
        & (25\%)Real + (75\%)Text2Pressure & $0.440 \pm 0.022$ & $0.515 \pm 0.021$ & $0.435 \pm 0.022$ & $0.465 \pm 0.019$ \\
        & (40\%)Real + (60\%)Text2Pressure & $0.475 \pm 0.021$ & $0.540 \pm 0.019$ & $0.460 \pm 0.023$ & $0.490 \pm 0.018$ \\
        & (50\%)Real + (50\%)Text2Pressure & $0.524 \pm 0.019$ & $0.582 \pm 0.022$ & $0.507 \pm 0.018$ & $0.513 \pm 0.020$ \\
        & (60\%)Real + (40\%)Text2Pressure & $0.480 \pm 0.020$ & $0.550 \pm 0.018$ & $0.495 \pm 0.020$ & $0.500 \pm 0.018$ \\
        & (75\%)Real + (25\%)Text2Pressure & $0.435 \pm 0.021$ & $0.525 \pm 0.019$ & $0.515 \pm 0.022$ & $0.475 \pm 0.017$ \\
        & (90\%)Real + (10\%)Text2Pressure & $0.420 \pm 0.022$ & $0.510 \pm 0.020$ & $0.530 \pm 0.021$ & $0.460 \pm 0.018$ \\
        & (100\%)Text2Pressure & $0.372 \pm 0.025$ & $0.463 \pm 0.022$ & $0.389 \pm 0.024$ & $0.421 \pm 0.019$ \\
        \hline
        \multirow{2}{*}{Pressure2Text} & (100\%)Real & $0.812 \pm 0.015$ & $0.832 \pm 0.016$ & 0.517 $\pm$ 0.014 & $0.801 \pm 0.019$ \\
        & (10\%)Real + (90\%)Text2Pressure & $0.790 \pm 0.017$ & $0.810 \pm 0.018$ & $0.400 \pm 0.016$ & $0.785 \pm 0.019$ \\
        & (25\%)Real + (75\%)Text2Pressure & $0.810 \pm 0.016$ & $0.825 \pm 0.017$ & $0.420 \pm 0.015$ & $0.800 \pm 0.018$ \\
        & (40\%)Real + (60\%)Text2Pressure & $0.830 \pm 0.015$ & $0.845 \pm 0.019$ & $0.440 \pm 0.014$ & $0.810 \pm 0.019$ \\
        & (50\%)Real + (50\%)Text2Pressure & 0.838 $\pm$ 0.013 & \textbf{0.861 $\pm$ 0.018} & $0.462 \pm 0.015$ & \textbf{0.819 $\pm$ 0.021} \\
        & (60\%)Real + (40\%)Text2Pressure & $0.835 \pm 0.014$ & $0.855 \pm 0.017$ & $0.470 \pm 0.014$ & $0.815 \pm 0.020$ \\
        & (75\%)Real + (25\%)Text2Pressure & $0.828 \pm 0.013$ & $0.850 \pm 0.018$ & $0.485 \pm 0.015$ & $0.810 \pm 0.019$ \\
        & (90\%)Real + (10\%)Text2Pressure & $0.820 \pm 0.014$ & $0.840 \pm 0.017$ & $0.500 \pm 0.015$ & $0.805 \pm 0.018$ \\
        & (100\%)Text2Pressure & $0.763 \pm 0.018$ & $0.690 \pm 0.019$ & $0.380 \pm 0.017$ & $0.671 \pm 0.020$ \\
        \hline
    \end{tabular}
    \label{tab:har_comparison_1_f1}
\end{table}

\begin{table}[!t]
\footnotesize
    \centering
    \caption{Comparison of Macro F1 across Different Datasets and Data Compositions for Pressure2Text over different LLM backbones.}
    \begin{tabular}{|c|c|c|c|c|c|}
        \hline
                \multirow{2}{*}{\textbf{Classifier}} & \multirow{2}{*}{\textbf{Training Set}} & \multicolumn{4}{c|}{\textbf{Macro F1} $\uparrow$}  \\ 
        \cline{3-6}
        & & \textbf{PresSim \cite{ray2023pressim}} & \textbf{TMD dataset} \cite{ray2024text} & \textbf{PmatData} \cite{pouyan2017pressure}& \textbf{MeX} \cite{wijekoon2019mex}\\
        \hline
        \multirow{2}{*}{} & (100\%)Real & $0.812 \pm 0.015$ & $0.832 \pm 0.016$ & \textbf{0.517 $\pm$ 0.014} & $0.801 \pm 0.019$ \\
        & (10\%)Real + (90\%)Text2Pressure & $0.790 \pm 0.017$ & $0.810 \pm 0.018$ & $0.400 \pm 0.016$ & $0.785 \pm 0.019$ \\
        & (25\%)Real + (75\%)Text2Pressure & $0.810 \pm 0.016$ & $0.825 \pm 0.017$ & $0.420 \pm 0.015$ & $0.800 \pm 0.018$ \\
        Pressure2Text& (40\%)Real + (60\%)Text2Pressure & $0.830 \pm 0.015$ & $0.845 \pm 0.019$ & $0.440 \pm 0.014$ & $0.810 \pm 0.019$ \\
        (Llama 2 13B)& (50\%)Real + (50\%)Text2Pressure & \textbf{0.838 $\pm$ 0.013} & \textbf{0.861 $\pm$ 0.018} & $0.462 \pm 0.015$ & \textbf{0.819 $\pm$ 0.021} \\
        & (60\%)Real + (40\%)Text2Pressure & $0.835 \pm 0.014$ & $0.855 \pm 0.017$ & $0.470 \pm 0.014$ & $0.815 \pm 0.020$ \\
        & (75\%)Real + (25\%)Text2Pressure & $0.828 \pm 0.013$ & $0.850 \pm 0.018$ & $0.485 \pm 0.015$ & $0.810 \pm 0.019$ \\
        & (90\%)Real + (10\%)Text2Pressure & $0.820 \pm 0.014$ & $0.840 \pm 0.017$ & $0.500 \pm 0.015$ & $0.805 \pm 0.018$ \\
        & (100\%)Text2Pressure & $0.763 \pm 0.018$ & $0.690 \pm 0.019$ & $0.380 \pm 0.017$ & $0.671 \pm 0.020$ \\
        \hline
        \multirow{2}{*}{} & (100\%)Real & $0.805 \pm 0.016$ & $0.825 \pm 0.017$ & \textbf{0.510 $\pm$ 0.013} & $0.795 \pm 0.018$ \\
        & (10\%)Real + (90\%)Text2Pressure & $0.781 \pm 0.018$ & $0.799 \pm 0.019$ & $0.392 \pm 0.017$ & $0.773 \pm 0.020$ \\
        & (25\%)Real + (75\%)Text2Pressure & $0.797 \pm 0.015$ & $0.814 \pm 0.018$ & $0.411 \pm 0.014$ & $0.789 \pm 0.017$ \\
        Pressure2Text& (40\%)Real + (60\%)Text2Pressure & $0.819 \pm 0.016$ & $0.836 \pm 0.020$ & $0.432 \pm 0.015$ & $0.802 \pm 0.018$ \\
        (Mistral 7B \cite{jiang2023mistral})& (50\%)Real + (50\%)Text2Pressure & \textbf{0.827 $\pm$ 0.014} & \textbf{0.851 $\pm$ 0.019} & $0.454 \pm 0.016$ & \textbf{0.811 $\pm$ 0.022} \\
        & (60\%)Real + (40\%)Text2Pressure & $0.824 \pm 0.015$ & $0.846 \pm 0.016$ & $0.461 \pm 0.015$ & $0.804 \pm 0.019$ \\
        & (75\%)Real + (25\%)Text2Pressure & $0.817 \pm 0.014$ & $0.839 \pm 0.017$ & $0.477 \pm 0.016$ & $0.798 \pm 0.018$ \\
        & (90\%)Real + (10\%)Text2Pressure & $0.809 \pm 0.015$ & $0.831 \pm 0.018$ & $0.491 \pm 0.014$ & $0.796 \pm 0.017$ \\
        & (100\%)Text2Pressure & $0.749 \pm 0.019$ & $0.678 \pm 0.020$ & $0.371 \pm 0.018$ & $0.659 \pm 0.021$ \\
        \hline
        \multirow{2}{*}{} & (100\%)Real & $0.798 \pm 0.016$ & $0.811 \pm 0.019$ & \textbf{0.496 $\pm$ 0.015} & $0.779 \pm 0.018$ \\
        & (10\%)Real + (90\%)Text2Pressure & $0.763 \pm 0.019$ & $0.783 \pm 0.020$ & $0.369 \pm 0.017$ & $0.742 \pm 0.021$ \\
        & (25\%)Real + (75\%)Text2Pressure & $0.774 \pm 0.018$ & $0.799 \pm 0.019$ & $0.387 \pm 0.016$ & $0.763 \pm 0.020$ \\
        Pressure2Text& (40\%)Real + (60\%)Text2Pressure & $0.786 \pm 0.017$ & $0.814 \pm 0.020$ & $0.402 \pm 0.016$ & $0.773 \pm 0.019$ \\
        (Gemma 2B \cite{team2024gemma})& (50\%)Real + (50\%)Text2Pressure & \textbf{0.791 $\pm$ 0.016} & \textbf{0.825 $\pm$ 0.019} & $0.429 \pm 0.017$ & \textbf{0.780 $\pm$ 0.022} \\
        & (60\%)Real + (40\%)Text2Pressure & $0.784 \pm 0.018$ & $0.819 \pm 0.019$ & $0.435 \pm 0.016$ & $0.773 \pm 0.021$ \\
        & (75\%)Real + (25\%)Text2Pressure & $0.775 \pm 0.016$ & $0.811 \pm 0.018$ & $0.445 \pm 0.016$ & $0.765 \pm 0.020$ \\
        & (90\%)Real + (10\%)Text2Pressure & $0.764 \pm 0.017$ & $0.803 \pm 0.018$ & $0.457 \pm 0.015$ & $0.758 \pm 0.019$ \\
        & (100\%)Text2Pressure & $0.715 \pm 0.020$ & $0.640 \pm 0.022$ & $0.342 \pm 0.018$ & $0.623 \pm 0.022$ \\
        \hline
    \end{tabular}
    \label{tab:har_comparison_2_f1}
\end{table}

\begin{table}[!t]
\footnotesize
    \centering
    \caption{Comparison of total number of parameters: computational cost and Floating Point Operations per Second(FLOPs): computational time for baseline classifiers, LLM-based classifier, and different Variations of Pressure2Text.}
    \begin{tabular}{|c|c|c|c|c|c|}
        \hline
        Classifier & LLM Backbone & Dataset &Parameters & FLOPs  \\    
        \hline
        Baseline & - & PresSim/TMD & 1.18$\times$10$^{6}$ & 0.00035$\times$10$^{12}$ \\
        Baseline & - & PmatData & 1.08$\times$10$^{6}$ & 0.00037$\times$10$^{12}$ \\
        Baseline & - & MeX & 0.33$\times$10$^{6}$ & 0.00105$\times$10$^{12}$ \\
        VideoLLaMA & Llama2 13B Chat  & PresSim/TMD/PmatData/MeX & 13386$\times$10$^{6}$ & 3.1$\times$10$^{12}$ \\
        Pressure2Text & Llama2 13B Chat & PresSim/TMD/PmatData/MeX & 13386$\times$10$^{6}$ & 3.3$\times$10$^{12}$ \\
        Pressure2Text & Mistral 7B  & PresSim/TMD/PmatData/MeX & 7306$\times$10$^{6}$ & 3.6$\times$10$^{12}$ \\
        Pressure2Text & Gemma 2B & PresSim/TMD/PmatData/MeX & 2512$\times$10$^{6}$ & 2.7$\times$10$^{12}$ \\
        \hline
    \end{tabular}
    \label{tab:computationalcost}
\end{table}
Pressure2Text offers a novel dynamic pressure classification mechanism. First, the dynamic pressure is converted to atomic motion descriptions. Then, we exploit the knowledge of LLM to classify the description into one target class. 
We created a classifier using prompt engineering within LLaMA 2 to map the sequence of atomic action descriptions into specific activity classes. For example, a description like \textit{a person bending down to his glutes} is processed through prompt engineering by framing it as a natural language prompt such as: \textit{Based on the description, classify the activity: {description}. Options: 0 (standing), 1 (walking), 2 (sitting), 3 (lying).} 
It assigns an integer corresponding to the most likely class, in this case, identifying the activity as sitting and outputting the integer \textit{2} that enables classification based on subtle contextual cues in the descriptions.
We evaluated the Pressure2Text classifier using the same four datasets with and without Text2Pressure data augmentation as shown in \cref{overview_eval}. 
Our results show improvement by 11.9$\%$ and 15.8$\%$ over the baseline for the TMD and MeX datasets, respectively, and an increase of 12.4$\%$ and 10$\%$ over the current state-of-the-art classifiers.

As stated in \cref{tab:har_comparison_1_f1}, the model trained with Text2Pressure-augmented data (with $50\%/50\%$ ratio) showed the best improvements across datasets, except PmatData where Text2Pressure generated the synthetic dataset and was not accurate, hence the decline in model accuracy.
Surprisingly, for the PresSim dataset, the Baseline classifier with Text2Pressure augmentation performed better than the Pressure2Text classifier, which can be attributed to the complexity of yoga motions when described as a sequence of atomic motions for training Cross-modal LLM-based classifiers.

\begin{figure}[!t]
\centering
\includegraphics[width=0.6\textwidth]{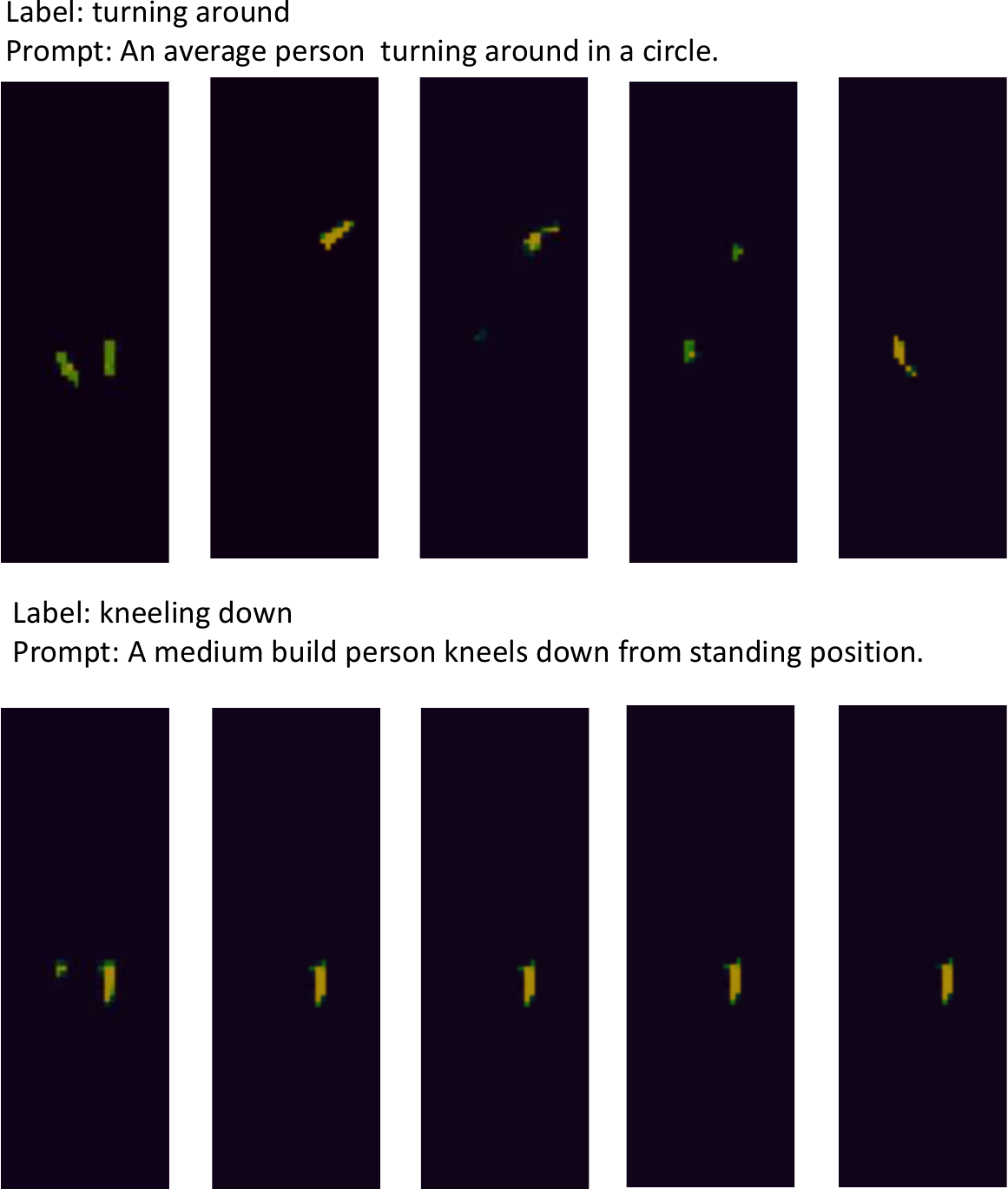} 
\caption{Figure depicting some instances of incorrect data generation by Text2Pressure.} 
\label{qual_limit2}
\end{figure}

\paragraph{LLM based Ablation}
Since Pressure2Text relies on an LLM at its core, and our experiments with LLaMA2 showed that a $50\%/50\%$ ratio performed best across most datasets, we aimed to assess whether changing the LLM would impact this outcome. We trained Pressure2Text using Mistral 7B and Gemma 2B as the backbone, as shown in \cref{tab:har_comparison_2_f1}. 
Our findings indicate that, regardless of the LLM used, the optimal combination remains $50\%/50\%$. LLMs with fewer parameters yield comparatively lower results than larger models, but they offer reduced computational costs, as detailed in \cref{tab:computationalcost}. 
Thus, users can select the most suitable LLM based on their specific computational and performance requirements.

\paragraph{Real-world Utility}
For Text2Pressure, speed and computational efficiency are not primary concerns since it is designed as a data generation model rather than a real-time classifier. However, compared to PresSim, which relies on physics simulations and a trained network, Text2Pressure is significantly faster. It can generate nearly 100 times faster synthetic pressure data, even before considering the additional time required to extract 3D pose information from video inputs (to prepare PresSim input). This makes Text2Pressure a far more efficient alternative for large-scale data generation, enabling rapid augmentation of pressure datasets without the computational overhead of physics-based simulations.

Although Pressure2Text, an LLM-based model, has significantly higher computational requirements than traditional classifiers due to its reliance on large language models, it is still highly efficient. As evident from the Floating Point Operations per Second (FLOPs) given in \cref{tab:computationalcost}, Pressure2Text operates much faster than its raw parameter count might suggest. It can run in real-time on an average graphics card like the NVIDIA RTX 3080 Mobile with minimal latency introduced by the VQ-VAE encoder. Furthermore, the model size can be reduced using quantized LLMs without significantly impacting performance, making deployment on edge devices more feasible. While its classification accuracy is not very high compared to traditional classifiers, Pressure2Text has the unique advantage of generating open-vocabulary text, allowing it to go beyond recognizing a fixed set of discrete classes and enabling more flexible and descriptive outputs.

\section{Discussion}
\label{sec:discussion}
\subsection{TxP for Wearable Pressure Sensors}

TxP is implemented on data generated by PresSim, which is designed to simulate pressure data from a pressure mattress. We require a PresSim-equivalent model that can generate wearable pressure sensor data from videos or 3D SMPL pose sequences to adapt this model for wearable pressure sensors, such as pressure insoles.

An alternative approach that allows us to use the existing PresSim weights is to modify its preprocessing steps. Pressure mattress dynamics and wearable pressure sensor data are interrelated. A pressure insole measures only the spatial pressure distribution of the foot concerning the ground, whereas a pressure mattress measures both the spatial pressure distribution of the foot and the position of the foot/body relative to the environment. If we can remove the environmental shift of the body from the generated pressure data, the resulting pressure dynamics would be equivalent to those of a pressure insole.

To achieve this, we modified the deformation profile simulation in PresSim by fixing the hip joint. Specifically, we set the rotation and position of the root joint (hip) to zero:

\begin{equation}
    \mathbf{p}_{\text{hip}} = (0, 0, 0), \quad \mathbf{R}_{\text{hip}} = I
\end{equation}

Since all other joints move relative to the root, this transformation preserves local motion while removing global body shifts. The deformation profiles generated from this modified simulation were then used to synthesize pressure data. During the simulation, two separate cameras are locked to only the left and right foot to remove the foot's relative motion relative to the body.

After generating synthetic pressure data with this approach, we applied the previously defined steps for mapping TxP to different pressure sensor configurations to ensure alignment with the spatial resolution of the wearable pressure sensors.

To evaluate the effectiveness of this approach, we trained TxP on the PID4TC dataset \cite{o2024ai}, a pressure insole dataset containing five activities: standing, walking, pick and place, assembly, and manual handling. Although the performance improvement with TxP was not as high as observed with the pressure mattress data, the results demonstrated potential as given in \cref{tab:insole_comparison}. The results are always worse than the baseline, especially when using the Pressure2Text classifier. This can be attributed to the fact that Pressure2Text is trained on pressure mattress data, which also has a spatial position of the pressure map relative to the environment. We cannot add that to the target dataset to make the data suitable for Pressure2Text, hence the bad performance. However, with a dedicated PresSim-equivalent model for wearable pressure simulation, we expect TxP to achieve even better results in wearable pressure sensing applications.

\begin{table}[!t]
\footnotesize
    \centering
    \caption{Comparison of HAR  on wearbale pressure insole data PID4TC \cite{o2024ai} with Text2Pressure and Pressure2Text Augementations.}
    \begin{tabular}{|c|c|c|}
        \hline
        \textbf{Classifier} & \textbf{Data} & \textbf{Macro F1 Score} $\uparrow$ \\ 
        \hline
        & (100\%)Real & 0.731 $\pm$ 0.018 \\
        & (10\%)Real + (90\%)Text2Pressure &  0.521 $\pm$ 0.024 \\
        & (25\%)Real + (75\%)Text2Pressure &  0.624 $\pm$ 0.018 \\
        & (40\%)Real + (60\%)Text2Pressure &  0.732 $\pm$ 0.014 \\
        Baseline & (50\%)Real + (50\%)Text2Pressure & \textbf{0.744} $\pm$ \textbf{0.035}\\
        & (60\%)Real + (40\%)Text2Pressure & 0.735 $\pm$ 0.019 \\
        & (75\%)Real + (25\%)Text2Pressure & 0.729 $\pm$ 0.018 \\
        & (90\%)Real + (10\%)Text2Pressure & 0.722 $\pm$ 0.023 \\
        & (100\%)Text2Pressure & 0.411 $\pm$ 0.012 \\
        \hline
        & (100\%)Real & 0.515 $\pm$ 0.018 \\
        & (10\%)Real + (90\%)Text2Pressure &  0.341 $\pm$ 0.022 \\  
        & (25\%)Real + (75\%)Text2Pressure &  0.432 $\pm$ 0.017 \\  
        & (40\%)Real + (60\%)Text2Pressure &  0.516 $\pm$ 0.015 \\  
        Pressure2Text & (50\%)Real + (50\%)Text2Pressure & 0.528 $\pm$0.029 \\  
        & (60\%)Real + (40\%)Text2Pressure & 0.521 $\pm$ 0.016 \\  
        & (75\%)Real + (25\%)Text2Pressure & 0.513 $\pm$ 0.018 \\  
        & (90\%)Real + (10\%)Text2Pressure & 0.502 $\pm$ 0.021 \\  
        & (100\%)Text2Pressure & 0.271 $\pm$ 0.013 \\ 
        \hline
    \end{tabular}
    \label{tab:insole_comparison}
\end{table}

\subsection{Application in other Sensor Domains}
Our approach to quantization is designed to handle long time-series data by segmenting it into smaller temporal chunks before applying vector quantization. This method has proven effective for data with rich spatial and temporal structures, such as pressure sensor arrays, depth sensors, and video sequences. However, many wearable sensors, such as IMUs, EMG, and bioimpedance sensors, primarily capture temporal information with sparse or no spatial structure, making direct quantization less effective. Some signals, like audio, which do not have much spatial information, can be converted to a format with rich spatial information, like Mel-spectrogram, to make it work with our architecture.

We propose a multi-modal quantization strategy that combines spatially rich sensor data with target sensor modalities to address this limitation. In this approach, we collect data from a modality with strong spatial-temporal characteristics (e.g., pressure, depth, or video) alongside data from the target sensor (e.g., IMU, EMG, or impedance). By quantizing both modalities together, we can leverage the structured spatial representations from the first modality to enhance the encoding of the target sensor's signal.

Additionally, we propose an alternative quantization method when sensor signals are simulated rather than recorded from real-world measurements. In cases where signals like IMU or EMG data are generated from 3D motion representations (e.g., SMPL pose sequences), we append the original 3D/SMPL pose data to the generated signals before quantization. This ensures that the quantized representation retains the structural constraints of the original motion, improving reconstruction and alignment with real-world sensor signals.

By incorporating these strategies, we can extend vector quantization's applicability beyond purely spatial-temporal data, enabling the effective encoding of diverse wearable sensor signals while maintaining consistency with natural sensor characteristics.

For some sensors, like IMUs, whose initial position and orientation also affect the generated signal, alternative strategies must be created to consider these parameters to generate a synthetic signal that can help improve HAR.

\subsection{Limitations}

\begin{figure}[!t]
\centering
\includegraphics[width=0.8\textwidth]{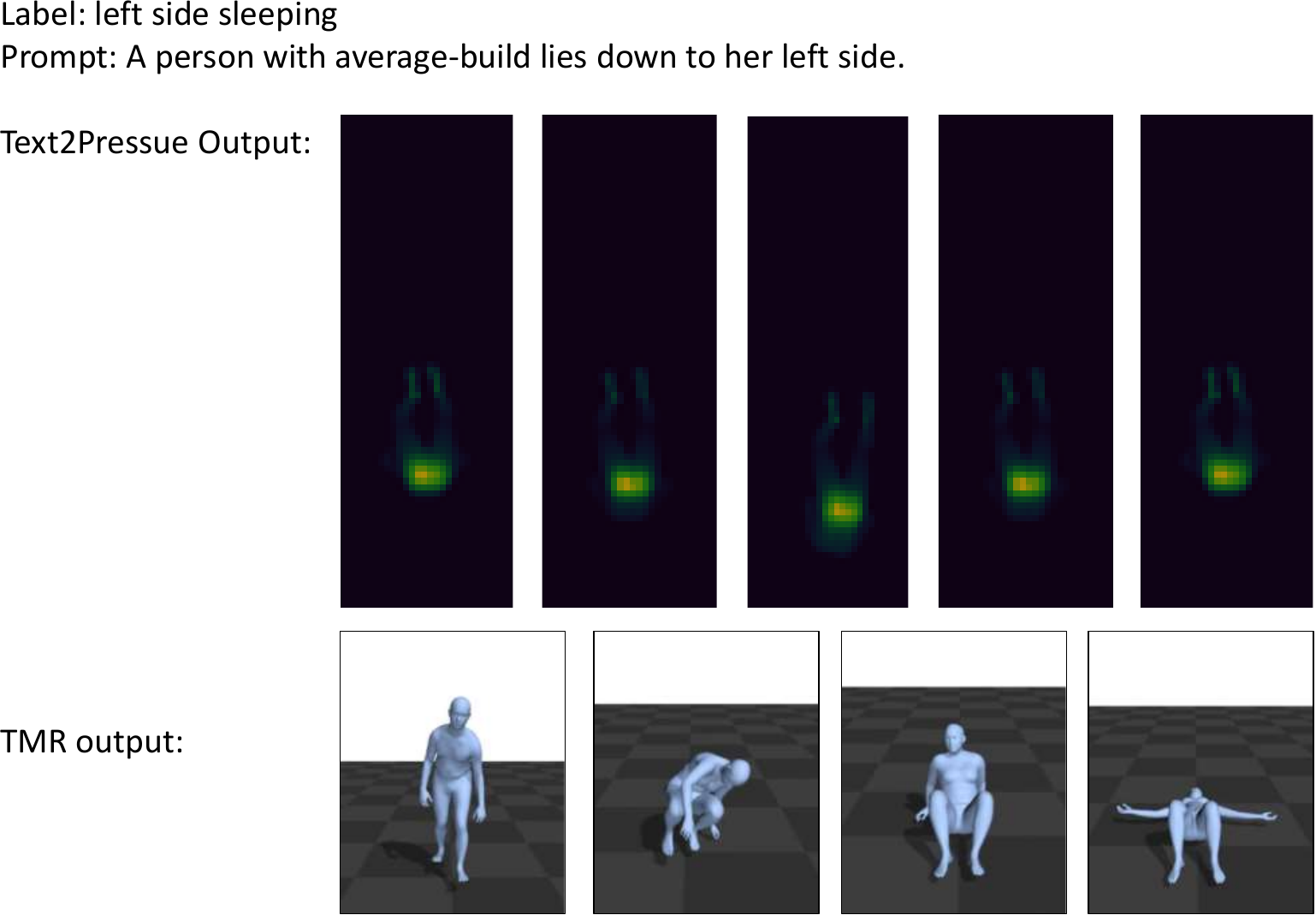} 
\caption{Figure depicting incorrect data generation by Text2Pressure for a particular activity class from PmatData, along with motion generated by the current SOTA text-to-motion model TMR having the same inaccuracy.} 
\label{qual_limit}
\end{figure}

Both Text2Pressure and Pressure2Text helped create a better pressure sensor-based HAR model, except for PmatData.
The performance of PmatData with data augmentation by TxP was notably worse compared to using only real data. 
This decline can be attributed to the original PressLang dataset, which needs more representations of lying-down motions. 
As the model is trained through motion quantization, any motion not included in the quantization process during the training of PressureRQVAE cannot be generated or recognized by Text2Pressure or Pressure2Text, respectively, as visualized in \cref{qual_limit2}. 

To address this limitation, we incorporated lying-down motions generated by the current SOTA text-to-motion model TMR \cite{petrovich2023tmr}. However, since these motions were trained on a subset of the motions used to create the PressLang dataset and employed a similar quantization technique with VQVAE, they failed to produce meaningful motions that could effectively expand and enhance TxP as visualized in \cref{qual_limit}.

We also noted that for complex activities such as yoga, which consist of multiple atomic actions in a specific sequence classified by the language model, Pressure2Text underperformed compared to a traditional classifier, as evidenced by the results on the PresSim dataset. This was surprising because Pressure2Text is more detailed and computationally more intensive.

This could be attributed to the inherent difficulty of capturing and contextualizing the intricate relationships between the various actions within a single quantized representation. Traditional classifiers operate on raw data streams and may handle complex motions more effectively. 
A better quantization process that considers the activity description needs to be adopted to solve this limitation. 

\subsection{LLM induced Biases}
While re-annotating PressLang dataset descriptions from Motion-X or generating synthetic activity descriptions from labels for downstream training, we did not implement methods to limit hallucinations by the LLMs or the generative models in TxP, and the bias can be seen in the evaluation score of PmatData. Approaches like IMUGPT 2.0 address this issue with a motion filter, and a similar classifier-based approach could be implemented to enhance the efficiency of the TxP framework. One solution for this bias resolution would be having a Pressure Dynamics filter akin to the Motion filter in IMUGPT 2.0, which is trained to verify if a pressure sequence/dynamics generated by TxP indeed matches the input text prompt through neural network-based classification into a discrete number of classes. If it is classified into the original class, it is qualified to be used as synthetic data; otherwise, it is rejected.

In the Context of Pressure2Text, translating hard labels to activity descriptions using LLM can introduce bias for certain activities, such as sleeping, yoga, or niche fitness exercises. Since LLMs are trained on general internet-scale text data, they may lack detailed descriptions for these activities. As a result, sleeping might receive overly simplistic annotations like "lying down," while yoga postures, which involve subtle variations and transitions, could be misrepresented or ignored entirely.
The best solution is to check the generated labels manually since each dataset contains 5-20 activity classes. An expert can easily verify them in a very short amount of time rather than relying on an overly complicated automated system.

Another significant issue is the fragmentation of complex activities into more conventional movements. Yoga, for instance, consists of continuous sequences of poses, but to fit HAR models like TxP, these might be artificially broken down into discrete actions such as "standing," "bending," or "stretching." This decomposition can cause TxP to lose the original temporal structure of yoga, leading to inconsistencies in generated pressure sequences. Similarly, activities that naturally last longer—like resting or meditation—may be forced into an unnatural segmentation, making it difficult for the model to recognize or generate them accurately. Again, bias like this can be overcome through manual checking, running the LLM inference multiple times, and taking the majority vote.

Additionally, bias evaluation metrics should be used to assess how different activities are represented in the dataset, ensuring that TxP does not disproportionately favor specific actions over others. We can improve TxP's ability to recognize and generate diverse human activities by addressing these biases, ultimately making HAR models more robust.

\section{Conclusion}
\label{sec:5}
In this paper, we introduce TxP, an innovative framework designed to transform the landscape of Human Activity Recognition (HAR) by generating realistic pressure sensor data from textual activity descriptions and synthesizing descriptive text from pressure sequences. TxP effectively addresses the critical issue of data scarcity in HAR, promoting interpretability and adaptability in models where pressure data is limited or absent.
Our bidirectional approach enhances HAR model performance in real-world sensor applications and significantly improves state-of-the-art benchmarks. Specifically, the Text2Pressure component augments data, boosting pressure-specific models by up to 11.6 \%. 
Furthermore, the Pressure2Text component introduces a novel classification strategy grounded in atomic motion patterns, advancing pressure-specific and multimodal language models by up to 12.4 \%.
With the addition of the PressLangDataset, TxP establishes a strong foundation for future research, supporting cross-modal learning and adaptable, efficient HAR models that meet real-world demands. 

Future efforts should incorporate unique activities, such as non-motion-based tasks like sleeping, critical for pressure sensing, to extend TxP's applicability to datasets like PmatData. Although Text2Pressure performed well across most cases, Pressure2Text was less effective for PmatData and PresSim datasets. TxP currently relies on PresSim for data generation, carrying over some of its imperfections; however, due to TxP's modular nature, PresSim can be easily replaced with a more accurate model as one becomes available. To further address this, adopting an improved token resampling approach or leveraging architectures like Mamba \cite{gu2023mamba}, which bypass the need for tokenization and work on continuous data streams, may yield better results for cross-modal training.
Another direction involves integrating other sensor modalities, such as IMU and EMG, to improve the robustness and adaptability of the TxP framework across diverse applications. Combining pressure sensing with complementary modalities could enhance activity recognition accuracy, particularly for complex or ambiguous movements that are challenging to capture with pressure sensors alone.

\begin{acks}
Multiple funding sources supported this research. It received funding from the German Federal Ministry of Education and Research (BMBF) under the VidGenSense project (01IW21003), as well as from the European Union's Horizon Europe research and innovation program under grant agreement No. 101162257 (STELEC). Additional support was provided by the Institute of Information \& Communications Technology Planning \& Evaluation (IITP), funded by the Korean government (MSIT), through the Artificial Intelligence Graduate School Program at Korea University (Grant No. RS-2019-II190079).
\end{acks}

\bibliographystyle{ACM-Reference-Format}
\bibliography{sample-base}
\appendix
\section{Codebook}
In a VQVAE, the codebook is a set of discrete latent embeddings that serve as learned prototypes for data representations. Unlike a standard Variational Autoencoder (VAE), which maps inputs to a continuous latent space with a Gaussian prior, VQ-VAE constrains each input representation to a finite set of quantized embeddings, making the distribution more structured and uniform.  
The codebook is learned during training through a discrete bottleneck mechanism consisting of the following steps: 
\begin{itemize}
    \item Encoding: The input is processed through an encoder, generating a continuous latent vector.
    \item Vector Quantization: Each latent vector is replaced by the nearest codebook entry, enforcing discrete representation.
    \item Commitment Loss: A penalty term ensures that the encoder consistently maps inputs to the duplicate codebook entries, preventing rapid switching.  
    \item Codebook Updates The embeddings in the codebook are updated using an exponential moving average (EMA) or the straight-through estimator to improve stability.
\end{itemize}  

Using a quantized codebook provides several benefits over standard VAEs by having a more uniform and limited latent distribution. Unlike VAEs, which may suffer from posterior collapse and excessive overlap in the latent space, VQ-VAE enforces well-separated, structured representations. Each codebook entry corresponds to a learned prototype, making latent representations more meaningful. The discrete latent space aligns well with tasks requiring structured representations, such as motion synthesis, sensor data modeling, and language generation, where a fixed set of latent tokens improves generalization and retrieval-based applications.

By enforcing a structured, quantized latent space, the VQ-VAE codebook enhances representation learning and generation quality, making it a powerful alternative to continuous latent models.  

\section{Residual Quantization in RQVAE}
The following content originates initially from RQ-VAE \cite{guo2024momask} and provides a comprehensive understanding of the residual quantization process within our framework.

\begin{equation}
b_v = Q(r_v), \quad r_{v+1} = r_v - b_v
\end{equation}

Here, $Q(r_v)$ represents the quantized output for the residual at layer $v$, and the residual is updated by subtracting the quantized value. This process iterates over $V+1$ quantization layers, yielding the final latent representation:

The quantized latent representation is then passed through a decoder $D$ to reconstruct the original pressure dynamics $\hat{p}$. Post-training, each pressure sequence is represented as $V+1$ discrete pressure token sequences:

\begin{equation}
T = [t^v_{1:n}]_{v=0}^{V}
\end{equation}

Each token sequence $T^v_{1:n} \in {1, \ldots, |C_v|}^n$ maps to the ordered codebook indices of the quantized embedding vectors $b^v_{1:n}$, where:

\begin{equation}
b^v_i = C^v t^v_i \quad \text{for} \quad i \in [1,n]
\end{equation}

We introduce a unique end token $V_x$ to mark sequence boundaries, appended after the last valid frame. This ensures alignment during sequence generation and reconstruction and clear termination indicators.

\paragraph{Loss Function}

The total loss function $L_{RQVAE}$ comprises a reconstruction loss and a quantization loss:

\begin{equation}
L_{RQVAE} = |p - \hat{p}|_1 + \sum{v=0}^{V} \beta_v |r_v - \text{sg}[b_v]|_2^2
\end{equation}

The reconstruction loss ensures accurate reconstruction, while the latent embedding loss encourages residuals to approximate their nearest codebook vectors. The stop-gradient operation $\text{sg}[\cdot]$ prevents backpropagation through the quantization step, with $\beta_v$ as a weighting factor.

\section{Text2Pressure Auto-regression}
The transformer model employs causal self-attention to predict the next index index autoregressively. The attention mechanism ensures that each index can only attend to its preceding indices, preventing information from future steps in the sequence from influencing the current token's prediction. This is formulated as:

\begin{equation}
\text{Attention} = \text{Softmax} \left( \frac{QK^T}{\sqrt{d_k}} \times \text{mask} \right)
\end{equation}

Where $ Q \in \mathbb{R}^{N \times d_k} $ and $ K \in \mathbb{R}^{N \times d_k} $ are the query and key matrices, and the mask is the causal mask that ensures that $ i $-th index cannot attend to future indices $ j > i $. During inference, starting from the text embedding, the generation proceeds autoregressively until the $ V_x $ is predicted, marking the end of the sequence.

\section{Text2Pressure Data Variations}  

\begin{figure}[!t]
\centering
\includegraphics[width=0.8\textwidth]{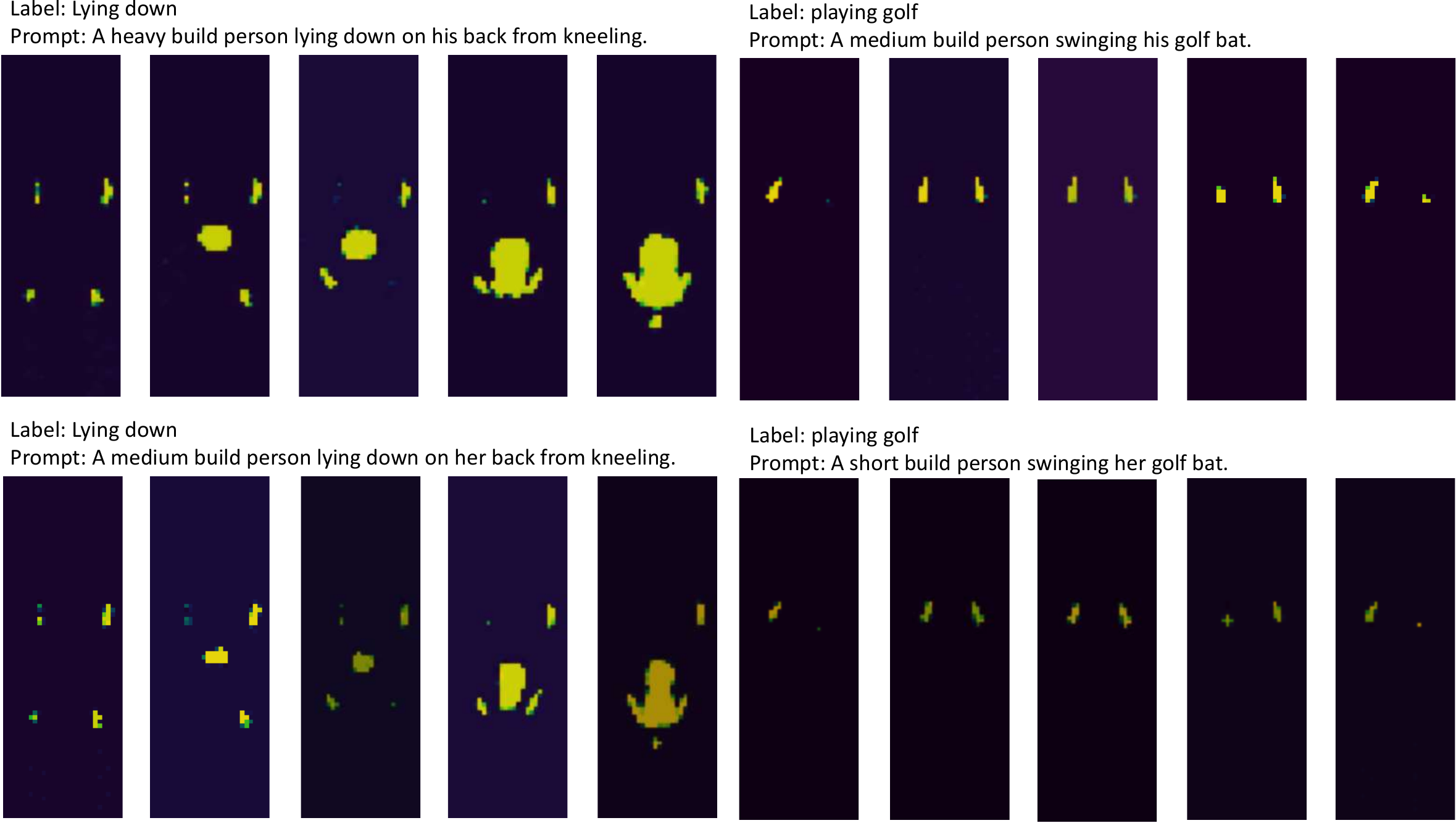} 
\caption{Figure depicting variations generated by Text2Pressure for an identical prompt about the activity, but changing the prompt for body type.} 
\label{variations}
\end{figure}

While generating pressure dynamics from SMPL motion sequences, we used each sequence five times, varying the 10 SMPL shape parameters and gender. To introduce body diversity, we randomly sampled SMPL bodies from the following six categories:  

Heavy tall male (\(80 \leq m < 100\) kg, \(\text{height} \geq 180\) cm), Medium tall male (\(60 \leq m < 80\) kg, \(\text{height} \geq 180\) cm), Light short male (\(40 \leq m < 60\) kg, \(\text{height} < 170\) cm), Heavy tall female (\(80 \leq m < 100\) kg, \(\text{height} \geq 170\) cm), Medium tall female (\(60 \leq m < 80\) kg, \(\text{height} \geq 170\) cm), Light short female (\(40 \leq m < 60\) kg, \(\text{height} < 160\) cm), where \(m\) denotes body mass and height is given in cm.

Our results indicated that the pressure dynamics of the light male and female SMPL bodies were nearly identical. Consequently, we merged them into a single category, reducing the total variations to five.

Additionally, we modified the soft labels and incorporated body information corresponding to the SMPL parameters used for generation. This adjustment enables our model to generate body-specific pressure dynamics, as demonstrated in \cref{variations}.  

For the Pressure2Text model, we removed gender-specific labels entirely and retained only body weight and height information\footnote{Following reviewer feedback and growing concerns in the literature regarding automatic gender recognition systems, we removed gender estimation capabilities from the Pressure2Text pipeline. While initial prototypes included gender information, we recognize that such features raise important ethical concerns around privacy, consent, and fairness \cite{hamidi2018gender, keyes2018misgendering}. Therefore, the model and associated outputs no longer consider gender, relying solely on body weight, height attributes, and activity.}.
\section{User guide for TxP for different Pressure sensor array sizes}
TxP is trained on simulated pressure data generated by PresSim, which in turn is trained to simulate data from a fitness mat developed by Sensing.Tex. The original sensor grid comprises \(80 \times 28\) pressure sensors, covering a sensing area of \(560 \times 1680\) mm. Since different applications have pressure sensor arrays of varying sizes, we used simple image processing techniques to adapt the simulated data to these configurations.

\begin{figure}[!t]
\centering
\includegraphics[width=0.8\textwidth]{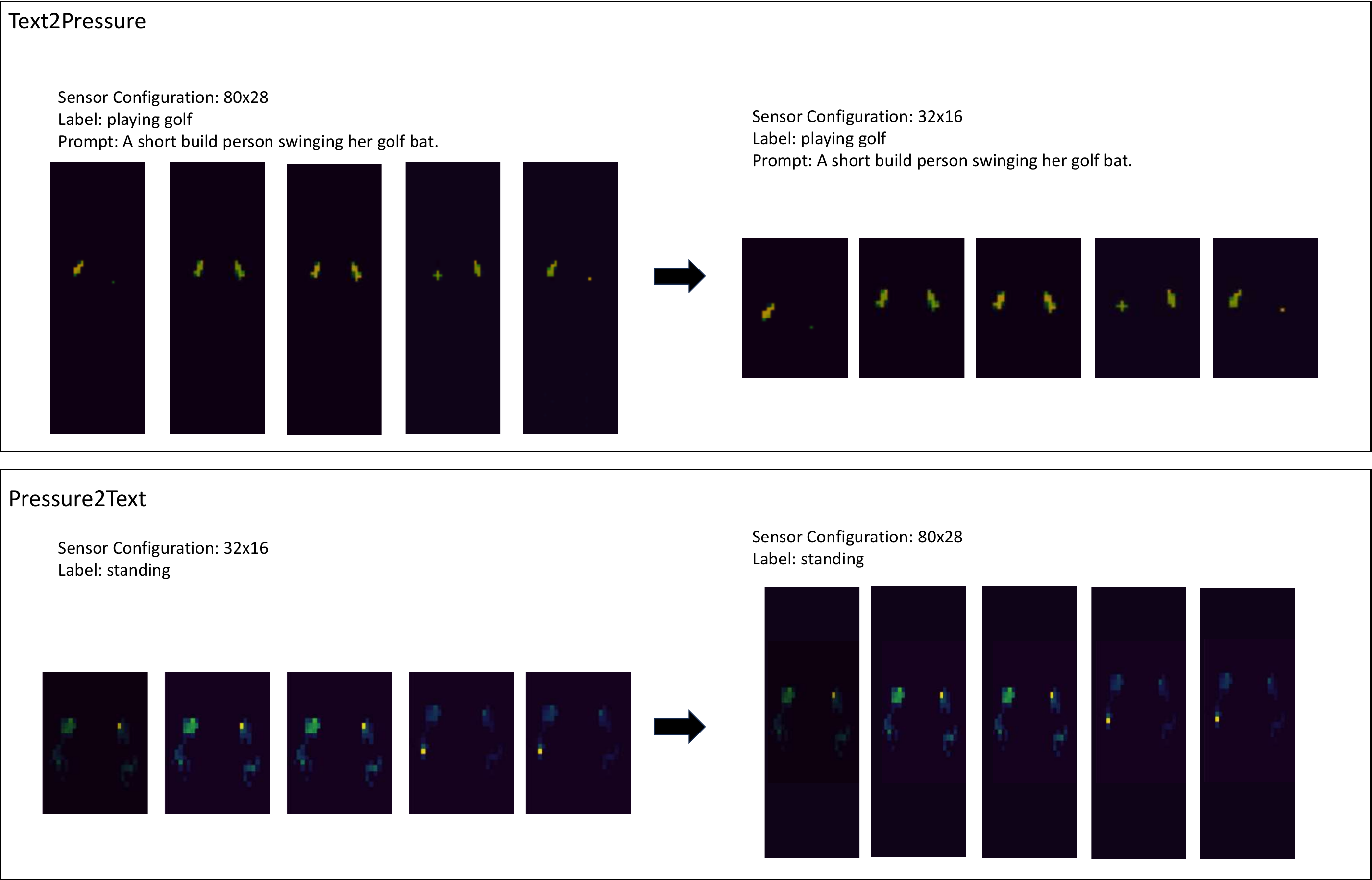} 
\caption{Figure depicting adaptation of TxP for different sensor configurations.} 
\label{adqaption}
\end{figure}
The original pressure grid is first centered and cropped to the desired size to match a target sensor array while maintaining the aspect ratio. Let:

\begin{itemize}
    \item \( P_{\text{orig}} \in \mathbb{R}^{560 \times 1680} \) be the original pressure grid.
    \item \( (c_x, c_y) \) be the center of \( P_{\text{orig}} \), computed as:
\end{itemize}

\begin{equation}
    c_x = \frac{560}{2}, \quad c_y = \frac{1680}{2}
\end{equation}

Define the cropping dimensions \( h_{\text{crop}}, w_{\text{crop}} \) based on the target sensor's aspect ratio. The cropped region is extracted as:

\begin{equation}
    P_{\text{crop}} = P_{\text{orig}}[c_x - h_{\text{crop}}/2 : c_x + h_{\text{crop}}/2, c_y - w_{\text{crop}}/2 : c_y + w_{\text{crop}}/2]
\end{equation}

Where \( h_{\text{crop}} \) and \( w_{\text{crop}} \) are chosen to match the target sensor's aspect ratio.

After cropping, the pressure grid is resampled to match the target sensor's resolution. If the target sensor grid has \( h_{\text{new}} \times w_{\text{new}} \) pressure sensors, bilinear interpolation is used:

\begin{equation}
    P_{\text{resampled}}(i, j) = P_{\text{crop}}\left( \frac{i}{h_{\text{new}}} h_{\text{crop}}, \frac{j}{w_{\text{new}}} w_{\text{crop}} \right)
\end{equation}

where \( P_{\text{resampled}} \in \mathbb{R}^{h_{\text{new}} \times w_{\text{new}}} \) is the final transformed pressure map.

Once \( P_{\text{resampled}} \) is obtained, it can be directly used as the output of Text2Pressure for target sensor-adapted simulated pressure dynamics. This transformation preserves the spatial structure of the pressure distribution, ensuring compatibility with the trained model while adapting to various sensor array sizes as shown in \cref{adqaption}. Similarly, we can use the inverse to fit any pressure data as input for Pressure2Text.

However, this method typically does not work for sensor arrays more extensive than the original \(560 \times 1680\) mm grid area, as the additional sensing area would contain no pressure spatial information. These regions would default to null values, which may not provide meaningful input to the model.

\end{document}